%% file: main.tex
\documentclass[10pt,twocolumn,letterpaper]{article}

\PassOptionsToPackage{table}{xcolor}
\usepackage{wacv}
\input{setup/package}
\input{setup/symbols}
\input{setup/math_commands}

\definecolor{wacvblue}{rgb}{0.21,0.49,0.74}
\usepackage[pagebackref,breaklinks,colorlinks,allcolors=wacvblue]{hyperref}

\def\wacvPaperID{396}
\def\confName{WACV}
\def\confYear{2027}

\title{{\modelAbbrev}: Load-Balanced and Efficient 3D Gaussian Splatting \\ for Large-Scale Scene Reconstruction}

\author{
Sheng-Hsiang Hung$^{1}$ \quad
Ting-Yu Yen$^{1}$ \quad
Wei-Fang Sun$^{2}$ \quad
Simon See$^{2}$ \\
Shih-Hsuan Hung$^{1}$ \quad
Hung-Kuo Chu$^{1}$\\[4pt]
$^{1}$National Tsing Hua University\quad
$^{2}$NVIDIA AI Technology Center (NVAITC)
}

\begin{document}
\maketitle
\input{0_Abstract}
\input{1_Introduction}
\input{2_RelatedWork}
\input{3_Method}
\input{4_Experiments}
\input{5_Conclusion}
\input{6_Acknowledgements}

{
    \small
    \bibliographystyle{ieeenat_fullname}
    \bibliography{main}
}

\clearpage
\appendix
\setcounter{figure}{0}
\setcounter{table}{0}
\setcounter{equation}{0}
\renewcommand{\thefigure}{A\arabic{figure}}
\renewcommand{\thetable}{A\arabic{table}}
\renewcommand{\theequation}{A\arabic{equation}}
\renewcommand{\theHfigure}{appendix.\arabic{figure}}
\renewcommand{\theHtable}{appendix.\arabic{table}}
\renewcommand{\theHequation}{appendix.\arabic{equation}}
\maketitlesupplementary

\input{7_Appendix}

\end{document}

%% file: setup/package.tex
\usepackage{bbm}
\usepackage{bm}
\usepackage{multirow}
\usepackage[accsupp]{axessibility}

\usepackage[ruled,vlined,linesnumbered]{algorithm2e}
\usepackage{float}
\usepackage{pifont}

%% file: setup/symbols.tex
\def\modelAbbrev{LoBE-GS}

\def\citygs{CityGS}

\def\vastgs{VastGS}

\def\dogs{DOGS}

\def\grendelgs{Grendel-GS}

\def\gsmod{3DGS$^\dagger$}

\newcommand{\Tfine}[1][]{%
  T_\mathrm{fine}^{#1}%
}

\def\fcs{\emph{fast camera selection}}

\definecolor{bestcolor}{RGB}{255,220,220}

\definecolor{secondcolor}{RGB}{255,235,205}

\definecolor{thirdcolor}{RGB}{255,255,210}

\definecolor{graycolor}{RGB}{224,224,224}

\newcommand{\best}[1]{\cellcolor{bestcolor}#1}

\newcommand{\second}[1]{\cellcolor{secondcolor}#1}

\newcommand{\third}[1]{\cellcolor{thirdcolor}#1}

\newcommand{\graybold}[1]{\cellcolor{graycolor}\textbf{#1}}

\DeclareRobustCommand{\bestlegend}[1]{%
  \begingroup
  \setlength{\fboxsep}{1pt}%
  \colorbox{bestcolor}{#1}%
  \endgroup
}

\DeclareRobustCommand{\secondlegend}[1]{%
  \begingroup
  \setlength{\fboxsep}{1pt}%
  \colorbox{secondcolor}{#1}%
  \endgroup
}

\DeclareRobustCommand{\thirdlegend}[1]{%
  \begingroup
  \setlength{\fboxsep}{1pt}%
  \colorbox{thirdcolor}{#1}%
  \endgroup
}

\DeclareRobustCommand{\graylegend}[1]{%
  \begingroup
  \setlength{\fboxsep}{1pt}%
  \colorbox{graycolor}{\textbf{#1}}%
  \endgroup
}

\newcommand{\cmark}{\ding{51}}
\newcommand{\xmark}{\ding{55}}

%% file: setup/math_commands.tex
\def\vp{{\bm{p}}}
\def\vu{{\bm{u}}}
\def\gB{{\mathcal{B}}}
\def\gC{{\mathcal{C}}}
\def\gN{{\mathcal{N}}}
\def\gP{{\mathcal{P}}}
\newcommand{\R}{\mathbb{R}}

%% file: 0_Abstract.tex
\begin{abstract}
3D Gaussian Splatting (3DGS) has established itself as an efficient representation for real-time, high-fidelity 3D scene reconstruction. However, scaling 3DGS to large and unbounded scenes such as city blocks remains difficult.
Existing divide-and-conquer methods alleviate memory pressure by partitioning the scene into blocks and training on multiple, non-communicating GPUs, but introduce new bottlenecks:
(i) partitions suffer from severe load imbalance since uniform or heuristic splits do not reflect actual computational demands, and
(ii) coarse-to-fine pipelines fail to exploit the coarse stage efficiently, often reloading the entire model and incurring high overhead.
In this work, we introduce {\modelAbbrev}, a \emph{\textbf{Lo}ad-\textbf{B}alanced and \textbf{E}fficient} 3D Gaussian Splatting framework, that re-engineers the large-scale 3DGS pipeline.
Specifically, {\modelAbbrev} introduces a load-balanced KD-tree scene partitioning scheme with optimized cutlines that balance per-block camera counts.
To accelerate preprocessing, it employs depth-based back-projection for fast camera assignment, reducing processing time from hours to minutes. It further reduces training cost through two lightweight techniques: visibility cropping and selective densification.
On large-scale urban and outdoor datasets, LoBE-GS reduces scene-partitioning time by up to an order of magnitude and worst-block fine-training time by up to $3.2\times$ relative to CityGaussian.
It also substantially reduces end-to-end and post-coarse training time under matched GPU budgets while maintaining competitive reconstruction quality.
LoBE-GS thus provides a communication-free alternative with favorable quality-efficiency trade-offs. The associated code is available at: \url{https://github.com/Sunnyhong0326/lobe-gs}
\end{abstract}

%% file: 1_Introduction.tex
\section{Introduction}
\label{sec:intro}

Recent advances in 3D scene reconstruction and novel view synthesis have shifted from photogrammetry and Neural Radiance Fields (NeRFs) toward explicit, real-time representations. Photogrammetry offers geometric precision but poor rendering efficiency, while NeRFs achieve photorealism at high computational cost. 3D Gaussian Splatting (3DGS) addresses both by representing scenes with millions of anisotropic Gaussian primitives under a GPU-friendly rasterization pipeline, delivering high fidelity at real-time speed, and has quickly become a leading representation for scalable 3D content creation.

Despite its success in bounded scenes, scaling 3DGS to large, unbounded environments such as city-scale reconstructions remains challenging: cost scales with the number of Gaussian primitives, making optimization time and GPU usage prohibitive. To mitigate this, recent works such as CityGaussian (\citygs)~\cite{liu2025citygaussian}, VastGaussian (\vastgs)~\cite{lin2024vastgaussian}, and DOGS~\cite{chen2024dogaussian} adopt a divide-and-conquer strategy, partitioning large scenes into spatial blocks processed in parallel. While this reduces memory pressure, it introduces new bottlenecks: current partitioning strategies lack \textbf{load balancing} because they do not account for computational load. Heuristics such as uniform grid splits or block-size normalization yield sub-regions with uneven optimization demands, so the slowest block dominates total training time, creating a long-tail bottleneck. \textbf{Coarse-to-fine pipelines remain inefficient} because methods such as \citygs~\cite{liu2025citygaussian} do not fully exploit the coarse stage to accelerate fine-level optimization, instead reloading the entire coarse model and incurring substantial overhead.

To overcome these limitations, we introduce \textbf{\modelAbbrev}, a framework that re-engineers the large-scale 3DGS pipeline for load-balanced, communication-free parallel training, addressing heuristic partitioning, coarse-model utilization, and fine-stage runtime.

A key bottleneck is data partitioning: existing methods evaluate all $M$ blocks against all $N$ cameras, requiring roughly $(M+1)\times N$ projections that can take hours. Instead, we compute a depth map once per camera from the coarse model, back-project it to 3D, and assign cameras via lightweight coverage tests, reducing preprocessing from hours to minutes.

To balance workload across blocks, we partition the scene with a load-balanced KD-tree that refines hierarchical cutlines to equalize predicted per-block workload, mitigating stragglers. We further reduce per-block cost with \emph{visibility cropping} and \emph{selective densification}, which prune irrelevant Gaussians and restrict densification while maintaining competitive rendering quality.

The main contributions of this work are summarized as follows:
\begin{itemize}
    \item We present \modelAbbrev, an efficient coarse-to-fine training framework for large-scale 3DGS, improving partition-based training through: (i) \textbf{load-balanced KD-tree partitioning} for adaptive scene decomposition; (ii) \textbf{fast camera selection} via depth-based back-projection to reduce partitioning overhead; and (iii) \textbf{visibility cropping and selective densification} to lower per-block cost while preserving quality.

    \item Experiments across large-scale benchmarks demonstrate that \modelAbbrev{} substantially reduces end-to-end and post-coarse training time under matched GPU budgets while maintaining competitive rendering quality.
\end{itemize}

%% file: 2_RelatedWork.tex
\begin{figure*}[t!]
\centering
\includegraphics[width=0.9\linewidth]{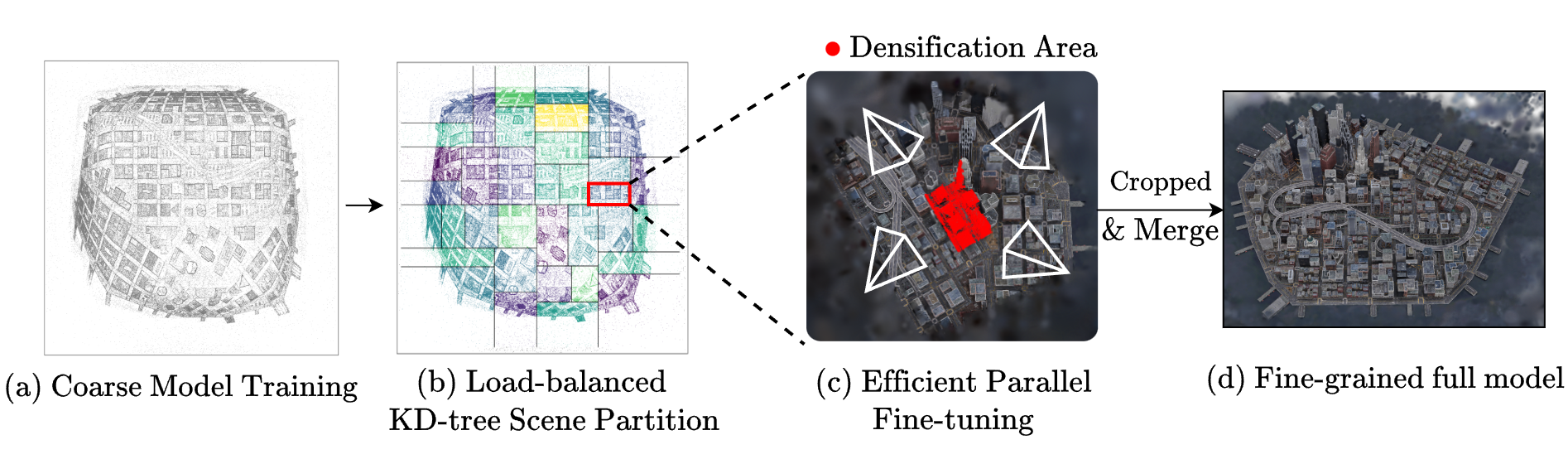}
\vspace{-1.5em}
\caption{\textbf{Overview of our framework.}
We first train a coarse 3DGS model. Next, we perform \textbf{load-balanced KD-tree partitioning} with \textbf{fast camera selection}, which partitions the scene into blocks and efficiently assigns high-coverage cameras to each block using reusable depth-based 3D cues. We further refine each split via \textbf{cutline refinement} to balance the workload across blocks. Before and during parallel fine-tuning, we apply \textbf{visibility cropping} and \textbf{selective densification} to reduce per-block optimization cost while preserving reconstruction quality. Finally, we prune Gaussians outside each block and merge all blocks into a unified, high-quality model.}
\label{pipeline}
\end{figure*}

\section{Related Work}
\label{sec:prior}

\subsection{Novel View Synthesis}
Novel view synthesis renders photorealistic scenes from unseen viewpoints. NeRF ~\cite{mildenhall2020nerf} models radiance fields with an MLP and volumetric integration, achieving high fidelity at substantial training and inference cost. 3D Gaussian Splatting (3DGS) ~\cite{kerbl3Dgaussians} instead represents scenes with Gaussian primitives and differentiable rasterization for efficient optimization and real-time rendering, but its quality degrades under sparse views and large scenes may require millions of primitives, motivating scalable reconstruction.

\subsection{Large-Scale Scene Reconstruction}
For decades, reconstructing large-scale 3D scenes has been a central goal~\cite{snavely2006photo, agarwal2011building}.
At city and regional scales, especially for aerial views~\cite{jiang2025horizon, tang2025dronesplat}, memory and throughput constraints motivate scalable training and rendering.

\noindent \textbf{Distributed training approaches}
jointly optimize a unified model across multiple GPUs.
NeRF-XL~\cite{li2024nerfxl} distributes NeRF training,
while \dogs~\cite{chen2024dogaussian} introduces Gaussian
consensus.
\grendelgs~\cite{zhao2024scaling3dgaussiansplatting}
distributes Gaussians using sparse all-to-all communication
and iterative load rebalancing, while
CityGS-X~\cite{Gao_2025_ICCV}
extends it with a shared decoder and cross-GPU synchronization.
Xie et al.~\cite{11047564} extend the Grendel framework with dynamic screen-space KD-tree partitioning to balance rendering workloads across GPUs and reduce communication overhead. In contrast, {\modelAbbrev} employs camera-count-guided KD-tree partitioning in 3D space to balance independently optimized blocks within a coarse-to-fine pipeline, further improving efficiency through fast camera selection, visibility cropping, and selective densification.
Unlike synchronized distributed training, {\modelAbbrev}
requires no inter-GPU communication during block optimization,
enabling flexible GPU scheduling and recovery.
We discuss interconnect sensitivity and deployment
considerations in Appendix~\ref{appendix:interconnect}.

\noindent \textbf{Partition-based approaches} partition a large scene into subregions, train submodels in parallel, and compose their outputs.
Block-NeRF~\cite{tancik2022block} assigns views to spatial blocks by camera position; Mega-NeRF~\cite{Turki2022CVPR} decomposes space into grids and routes rays to intersected grids; Switch-NeRF~\cite{mi2023switchnerf} learns decomposition and routing end-to-end via a mixture-of-experts.
Within 3DGS, {\vastgs}~\cite{lin2024vastgaussian} introduces progressive spatial partitioning and assigns cameras/points using an \textit{airspace}-aware visibility criterion, then fuses independently optimized blocks.
{\citygs}~\cite{liu2025citygaussian} further leverages a coarse global 3DGS prior to guide training and fusion, and contracts unbounded scenes to a normalized bounded cubic space before uniform grid partitioning.
However, for parallel scalability, uniform partitioning can still yield stragglers when submodels have imbalanced visibility workloads.
Moreover, coarse-prior guided pipelines may incur non-trivial overhead when the prior is large and must be accessed across many workers.
To address these, {\modelAbbrev} balances the coarse 3DGS prior across submodels within each subregion and trains them efficiently in parallel.

\subsection{Efficient Gaussian Splatting Reconstruction}
Beyond scalability, many works improve the efficiency of 3DGS reconstruction and rendering.
3DGS compression~\cite{navaneet2023compact3d, papantonakisReduced3DGS} reduces storage, while Taming 3DGS~\cite{mallick2024taming} addresses budget-constrained training via guided densification toward high-contribution Gaussians.
For large-scale scenes, level-of-detail (LoD) 3DGS representations enable more efficient rendering~\cite{ren2024octree, hierarchicalgaussians24}.
CityGaussianV2~\cite{liu2024citygaussianv2} builds on {\citygs}~\cite{liu2025citygaussian} and incorporates 2DGS for improved geometric modeling.
Momentum-GS~\cite{Fan_2025_ICCV} extends Scaffold-GS~\cite{scaffoldgs} with momentum self-distillation and reconstruction-guided block weighting to improve large-scene parallel training.
In this work, we focus on efficient large-scale 3DGS reconstruction with a coarse 3DGS prior and load-balanced, communication-free parallel training on multiple GPUs.

%% file: 3_Method.tex
\section{Methodology}
\label{sec:method}

Prior large-scale 3DGS systems achieve strong reconstruction quality, but their training efficiency is often limited by workload imbalance across blocks, expensive camera assignment, and unnecessary optimization of irrelevant Gaussians during block-wise fine-training. To address these issues, we propose {\modelAbbrev}, a coarse-to-fine training framework that improves the efficiency of partition-based 3DGS training while preserving reconstruction quality. As illustrated in Figure~\ref{pipeline}, our framework consists of three key components: (1) {\emph{load-balanced KD-tree scene partition}} (Sec.~\ref{sec:methodology:load-balance-aware_scene_partition}); (2) {\fcs} (Sec.~\ref{sec:methodology:fast_camera_selection}) and (3) {\emph{visibility cropping}} and {\emph{selective densification}} (Sec.~\ref{sec:methodology:visibility_cropping_and_selective_densification}).

\subsection{Load-balanced KD-tree Scene Partition}
\label{sec:methodology:load-balance-aware_scene_partition}
To enable efficient parallel fine-tuning, we partition the scene into $M$ spatial
blocks so that each block can be trained on a GPU. We use a KD-tree, where leaves
correspond to the final blocks and each node stores an axis-aligned bounding box
(AABB) $\gB^{(b)}$ and an associated camera set $\gC^{(b)}$. We define a lightweight
workload estimate of a node as its camera count, $L(\gB^{(b)}) = |\gC^{(b)}|$, a
proxy we validate against alternatives in
Sec.~\ref{sec:experiments:proxy_analysis}.

\noindent \textbf{KD-tree construction.} Starting from the global scene bounds, we recursively split nodes into two children by a cut plane orthogonal to one axis.
For a node, we choose the split axis $a \in \{x,y\}$ using a geometric heuristic (e.g., the longest AABB extent) to avoid degenerate thin blocks and to maintain spatial compactness.
We initialize the cut position along axis $a$ (e.g., the median of the coarse Gaussian centers in $\gB^{(b)}$ projected onto the split axis) to obtain a reasonable spatial separation before balancing.
To reduce the global maximum workload, we grow the tree in a best-first manner: at each step we split the current leaf with the largest $L(\cdot)$.
This strategy prioritizes splitting the current bottleneck leaf, directly targeting the global maximum workload rather than expanding the tree along a fixed traversal order. The KD-tree can also terminate on a per-node camera budget, so scenes beyond single-model capacity set $N_\text{coarse} > 1$  and train coarse blocks in parallel (Appendix~\ref{appendix:implementation_details}).

\input{figure_tex/cutline_refinement_analysis}

\noindent \textbf{Cutline refinement.} Although KD-tree partitioning provides a spatially
compact hierarchical decomposition, a naive split location (e.g., \ median split) may still
lead to substantial workload imbalance between sibling nodes. We therefore refine each
split by adjusting the cutline along the selected axis. Let \(p \in [0,1]\) be a split
percentile over the coarse Gaussian position projected onto the chosen axis, and let
\(s(p)\) denote the corresponding spatial cutline coordinate, i.e., \ the \(p\)-th quantile
of those projected centers. The cutline \(s(p)\) partitions the parent node into the two
children \(\gB_L(s(p))\) and \(\gB_R(s(p))\), and we define the objective
\begin{equation}
J(p) = \max\bigl(L(\gB_L(s(p))),\; L(\gB_R(s(p)))\bigr),
\end{equation}
where \(L(\cdot)\) denotes the workload of a region. The goal is to find
\begin{equation}
p^\ast \;=\; \operatorname*{arg\,min}_{p \in [p_{\min},\, p_{\max}]} \; J(p),
\qquad
s^\ast = s(p^\ast),
\end{equation}
where $[p_{\min}, p_{\max}]$ prevents degenerate splits. We set $p_{\min}=0.1$ and $p_{\max}=0.9$. Searching over $p$ rather than the raw coordinate makes the search range scene-independent, while the monotonic mapping $s(\cdot)$ preserves the spatial ordering of candidate splits. The minimax objective aims to balance the camera-count workloads of the two child blocks.

Figure~\ref{fig:cutline_refinement} illustrates the opposing camera-count trends as the cutline moves. We approximately minimize the maximum sibling workload using a lightweight binary-search procedure. At each iteration, we shift the search interval toward the more heavily loaded child and retain the candidate with the lowest maximum camera count. This provides an efficient approximation to the minimax objective without exhaustively evaluating all possible cutlines.

\subsection{Fast Camera Selection}
\label{sec:methodology:fast_camera_selection}

\emph{Camera selection} assigns each block a subset of cameras $\gC^{(b)}$, discarding views with negligible coverage of the block so that each block is optimized only with relevant views.

Despite its importance, prior studies often overlook the computational burden of this process, which can dominate the partitioning stage and, in coarse-to-fine pipelines, account for a substantial fraction of preprocessing time. For instance, CityGS computes per-block SSIM against the full coarse render, requiring at least (M + 1) × N renders and dominating partitioning cost (Sec.~\ref{sec:experiments:load_balance_and_runtime}).

To eliminate this overhead, we introduce \fcs{}, which amortizes the expensive 3D-to-2D projection by performing a \emph{single} depth compositing pass per camera view (i.e., $|\mathcal{C}|$ renders in total) and reusing the resulting back-projected points throughout KD-tree partitioning and refinement. For each camera view, we compute an opacity-normalized depth map. For a camera \(c\) and pixel \(\vu\), let \(\gN_{c,\vu}\) be the ordered set of points along the corresponding ray, and let \(w_{i,c,\vu}=\alpha_{i,c,\vu}\prod_{j=1}^{i-1}(1-\alpha_{j,c,\vu})\)
be the alpha-compositing weight of the \(i\)-th point, where \(d_{i,c}\) is its depth in the
camera frame and \(\alpha_{i,c,\vu}\) is determined by covariance and opacity. The
accumulated opacity and normalized depth at that pixel are
\begin{equation}
A_c(\vu)=\sum_{i=1}^{|\gN_{c,\vu}|} w_{i,c,\vu},~
D_c(\vu)=\frac{\sum_{i=1}^{|\gN_{c,\vu}|} w_{i,c,\vu}\, d_{i,c}}{A_c(\vu)+\epsilon}.
\end{equation}
Back-projecting every pixel of the depth map into 3D space yields a dense point cloud
\begin{equation}
\gP^{(c)}=\{\vp_{c,k}\mid k=1,\ldots,N_c\},~
\vp_{c,k}\in\R^3
\end{equation}
where \(N_c\) is the number of pixels in view \(c\). Next, for each camera \(c\) and block \(b\), we compute the visibility ratio
of points inside the block:
\begin{equation}
V_{c,b} = \frac{1}{N_c} \sum_{k=1}^{N_c} \mathbbm{1}[\vp_{c,k} \in \gB^{(b)}],
\end{equation}
where \(\mathbbm{1}\) denotes the indicator function, and \(\gB^{(b)}\) is the spatial region
of block \(b\). We apply no opacity-based filtering during back-projection and normalize block coverage by the total image pixel count. Despite this approximation, our camera selection shows strong agreement with the CityGS reference (Appendix~Table~\ref{tbl:camera_agreement}). The assigned camera set for block $b$ is defined as $\gC^{(b)}=\{ c \mid V_{c,b} \ge \tau \}$, where $\tau=0.1$ is a threshold to prune views with negligible block coverage. Moreover, during KD-tree splitting we only redistribute the cameras already assigned to the parent block (instead of re-evaluating all cameras), so the cost of each split scales with the parent camera set size, enabling efficient recursive partitioning and cutline refinement.

\input{table_tex/quantitative_results}
\input{table_tex/runtime_quality}
\subsection{Visibility Cropping and Selective Densification}
\label{sec:methodology:visibility_cropping_and_selective_densification}
Prior coarse-to-fine 3DGS pipelines load the entire coarse model during per-block fine-training, incurring memory and runtime overhead.
The memory overhead comes from storing all coarse Gaussians in GPU memory.
The runtime overhead comes mainly from the Adam optimizer rather than rendering: frustum culling already excludes non-visible Gaussians during rasterization, but Adam still maintains states and updates parameters for all loaded Gaussians, including those unobserved by the block's cameras~\cite{mallick2024taming}.

A straightforward fix is to keep only the Gaussians whose centers lie inside the current block, but this naive spatial cropping degrades quality, since some Gaussians outside the block remain visible in its assigned views and are still needed during fine-training.
We therefore introduce \emph{visibility cropping}: for each
block, we retain only Gaussians whose projected footprints
are visible in at least one assigned camera view, rather than
loading the entire coarse model. This reduces the number
of optimized Gaussians while preserving relevant scene content
for fine-training.

Visibility cropping may still retain Gaussians outside the block boundary when they are visible in the assigned views.
Such Gaussians provide useful rendering supervision but are discarded before block merging, so allowing them to densify is unnecessary.
Motivated by this, we introduce \emph{selective densification}, which restricts densification to Gaussians inside the current block.
Outside Gaussians thus still contribute to rendering and optimization when needed, but do not spawn new Gaussians that will later be removed, reducing memory usage and improving optimization efficiency while preserving per-block fidelity.

%% file: figure_tex/cutline_refinement_analysis.tex
\begin{figure}[t]
    \centering

    \begin{subfigure}[t]{0.47\linewidth}
        \centering
        \includegraphics[width=\linewidth]{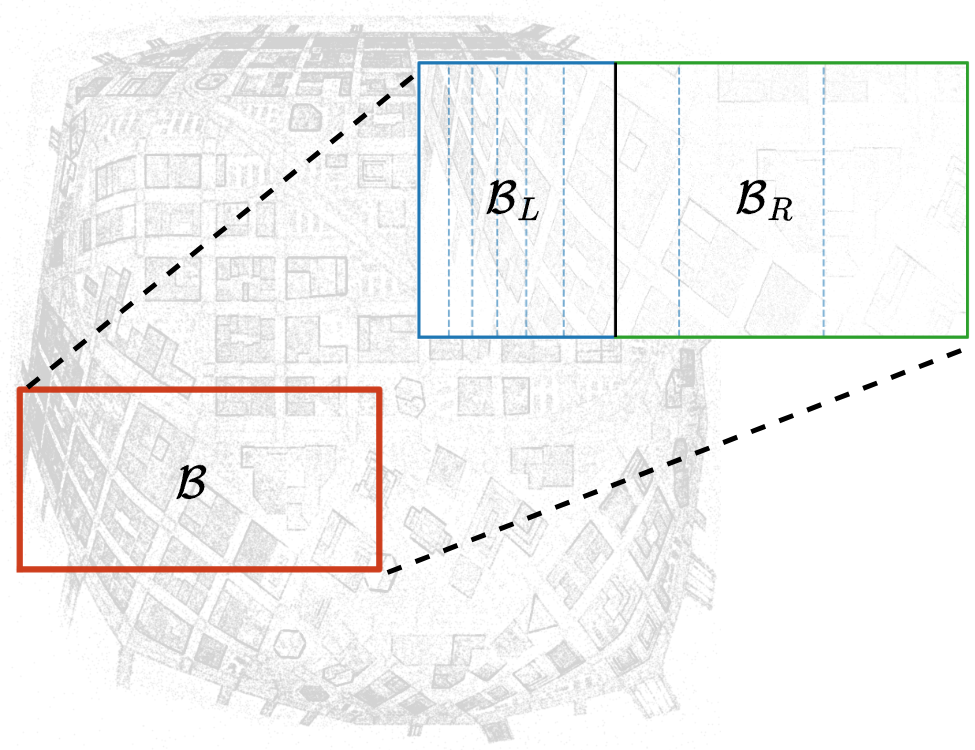}
        \captionsetup[subfigure]{justification=centering}
        \caption{\small{Parent AABB and cutline candidates}}
        \label{fig:cutline_refinement_left}
    \end{subfigure}
    \hfill
    \begin{subfigure}[t]{0.5\linewidth}
        \centering
        \includegraphics[width=\linewidth]{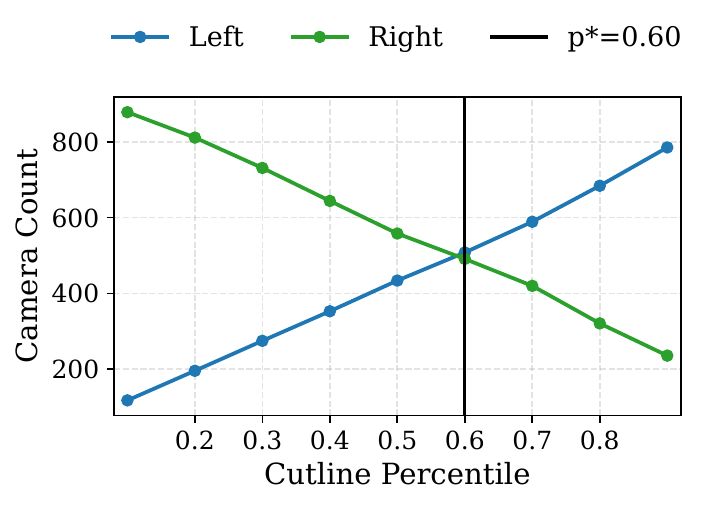}
        \captionsetup[subfigure]{justification=centering}
        \caption{\small{Sibling camera count vs. cutline percentile}}
        \label{fig:cutline_refinement_right}
    \end{subfigure}
    \vspace{-0.8em}
    \caption{\textbf{Cutline refinement in load-balanced KD-tree scene partition.}
    (a) A parent AABB is selected for splitting, and multiple cutline candidates are evaluated along the chosen axis in the zoomed region.
    (b) As the cutline percentile increases, the left-child camera count increases while the right-child camera count decreases, producing an approximately monotonic trade-off.
    We approximately minimize the maximum sibling camera count via binary search.}
    \label{fig:cutline_refinement}
\end{figure}

%% file: table_tex/quantitative_results.tex
\begin{table*}[t!]
\centering
\renewcommand{\arraystretch}{1.1}
\setlength{\tabcolsep}{4pt}
\resizebox{\textwidth}{!}{%
\begin{tabular}{l|cccr|cccr|cccr|cccr|cccr}
\toprule
\multirow{2}{*}{Method} &
\multicolumn{4}{c|}{MatrixCity-BigCity} &
\multicolumn{4}{c|}{MatrixCity-SmallCity} &
\multicolumn{4}{c|}{Building} &
\multicolumn{4}{c|}{Rubble} &
\multicolumn{4}{c}{Residence} \\
& PSNR & SSIM & LPIPS & $T_{\text{E2E}}^{(8)}$
& PSNR & SSIM & LPIPS & $T_{\text{E2E}}^{(8)}$
& PSNR & SSIM & LPIPS & $T_{\text{E2E}}^{(8)}$
& PSNR & SSIM & LPIPS & $T_{\text{E2E}}^{(8)}$
& PSNR & SSIM & LPIPS & $T_{\text{E2E}}^{(8)}$ \\
\midrule
\multicolumn{21}{l}{\textbf{Single-model / distributed methods}} \\
{\gsmod}
& 22.64 & 0.660 & 0.454 & 24:29
& 25.57 & 0.794 & 0.315 & 02:10
& 20.46 & 0.720 & 0.305 & 01:30
& 25.47 & \third{0.777} & \third{0.277} & 01:10
& 21.44 & 0.791 & 0.236 & 01:22 \\
\grendelgs
& 23.01 & 0.740 & 0.354 & 10:46
& \third{26.96} & \third{0.850} & \third{0.224} & 02:05
& \best{22.52} & \third{0.764} & \third{0.277} & 00:46
& \third{25.72} & 0.768 & 0.301 & 00:30
& \best{22.37} & \third{0.804} & \third{0.227} & 00:43 \\
\midrule
\multicolumn{21}{l}{\textbf{Partition-based methods}} \\
\vastgs
& \best{29.17} & \best{0.889} & \second{0.173} & 57:29
& 26.92 & 0.845 & 0.235 & 04:31
& 21.31 & 0.743 & 0.277 & 02:28
& 25.13 & 0.749 & 0.316 & 01:15
& 20.88 & 0.778 & 0.242 & 02:28 \\
\citygs
& \third{27.87} & \third{0.838} & \third{0.265} & 140:31
& \second{27.46} & \second{0.865} & \second{0.204} & 06:53
& \third{21.55} & \second{0.778} & \second{0.246} & 04:51
& \second{25.77} & \best{0.813} & \best{0.228} & 03:33
& \second{22.00} & \best{0.813} & \second{0.211} & 06:11 \\
{\modelAbbrev}
& \second{28.88} & \second{0.888} & \best{0.170} & \graybold{15:27}
& \best{27.69} & \best{0.875} & \best{0.185} & \graybold{02:08}
& \second{22.13} & \best{0.790} & \best{0.231} & \graybold{01:33}
& \best{25.93} & \best{0.813} & \second{0.232} & \graybold{01:01}
& \third{21.48} & \second{0.805} & \best{0.209} & \graybold{01:38} \\
\bottomrule
\end{tabular}}
\vspace{-2mm}
\caption{\textbf{Quantitative comparison on MatrixCity, Mill19, and UrbanScene3D datasets.}
We report PSNR, SSIM, LPIPS, and end-to-end runtime $T_{\text{E2E}}^{(8)}$ under an 8-GPU budget.
The \bestlegend{best}, \secondlegend{second}, and \thirdlegend{third} best reconstruction quality are highlighted across five methods.
For runtime, the fastest method among partition-based methods is highlighted in \graylegend{bold}.
All runtimes are reported in $hh$:$mm$.
Full quantitative details are reported in Appendix~Table~\ref{tbl:quantitative_full}.}
\label{tbl:quantitative_comparison}
\end{table*}

%% file: table_tex/runtime_quality.tex
\begin{figure*}[!htbp]
\centering
\includegraphics[width=0.5\textwidth]{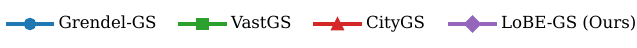}

\vspace{-0.5em}
\begin{subfigure}[t]{0.2\textwidth}
    \centering
    \includegraphics[width=\linewidth]{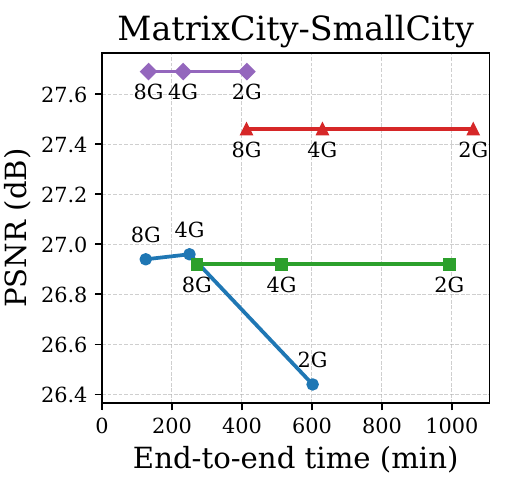}
    \caption{MC-SmallCity, PSNR}
    \label{fig:rqt_matrixcity_psnr}
\end{subfigure}
\hfill
\begin{subfigure}[t]{0.2\textwidth}
    \centering
    \includegraphics[width=\linewidth]{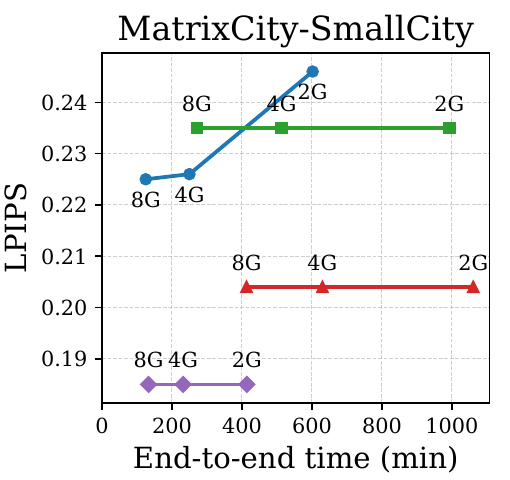}
    \caption{MC-SmallCity, LPIPS}
    \label{fig:rqt_matrixcity_lpips}
\end{subfigure}
\hfill
\begin{subfigure}[t]{0.2\textwidth}
    \centering
    \includegraphics[width=\linewidth]{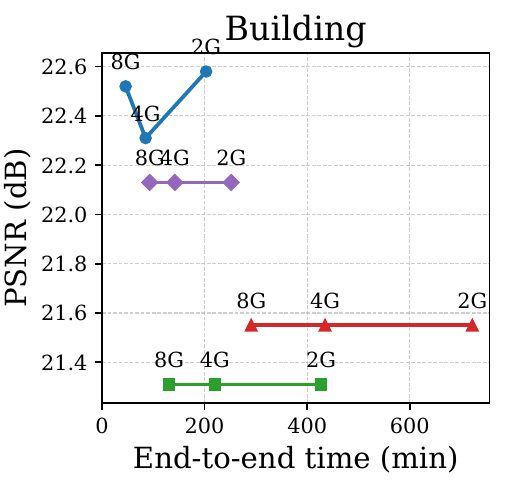}
    \caption{Building, PSNR}
    \label{fig:rqt_building_psnr}
\end{subfigure}
\hfill
\begin{subfigure}[t]{0.2\textwidth}
    \centering
    \includegraphics[width=\linewidth]{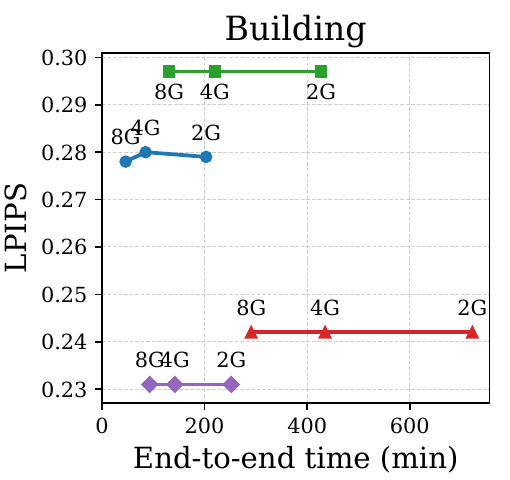}
    \caption{Building, LPIPS}
    \label{fig:rqt_building_lpips}
\end{subfigure}
\vspace{-0.7em}
\caption{\textbf{Runtime-quality trade-off under matched GPU budgets.}
We compare reconstruction quality and end-to-end runtime under 2, 4, and 8 GPUs across two datasets and two metrics: (a,b) MatrixCity-SmallCity and (c,d) Building, with PSNR and LPIPS reported per dataset. For PSNR, upper-left is better; for LPIPS, lower-left is better. Markers indicate GPU budgets, and lines connect runs of the same method. The full runtime-quality trade-off is reported in Appendix~Fig.~\ref{fig:appendix_runtime_quality_tradeoff}.}
\label{fig:runtime_quality_tradeoff}
\end{figure*}

%% file: 4_Experiments.tex
\section{Experiments}
\label{sec:experiments}

\textbf{Datasets.} We conducted experiments on six large-scale scenes from three datasets, including four real-world scenes and two synthetic scenes. For the real-world datasets, we used \emph{Building} and \emph{Rubble} from Mill19~\cite{Turki2022CVPR}, and \emph{Residence} and \emph{Sci-Art} from UrbanScene3D~\cite{UrbanScene3D}. For the synthetic dataset, we adopted \emph{MatrixCity-BigCity} and \emph{MatrixCity-SmallCity}~\cite{li2023matrixcity}.
We present quantitative comparisons on five scenes in the main paper, with additional results for Sci-Art provided in Appendix~\ref{appendix:additional_quantitative_results}.
Following prior work~\cite{liu2025citygaussian}, all images in MatrixCity were resized to a width of 1600 pixels.
For a fair comparison on real-world datasets, we downsampled all images by a factor of four, consistent with previous methods.

\noindent \textbf{Baselines.} We compare against state-of-the-art large-scale 3DGS methods, including {\citygs}~\cite{liu2025citygaussian}, {\vastgs}~\cite{lin2024vastgaussian}, and {\grendelgs}~\cite{zhao2024scaling3dgaussiansplatting}. We also include {\gsmod}, which follows the original 3DGS pipeline but trains for 60k iterations with densification every 200 iterations until 30k iterations. Since the official VastGS implementation is not publicly available, we use an unofficial implementation\footnote{\url{https://github.com/kangpeilun/VastGaussian}} without appearance modeling. We additionally apply point-cloud filtering on MatrixCity; Appendix~Table~\ref{tbl:vastgs_reproduction} reports the impact of this modification. For {\grendelgs}, we use the official implementation with COLMAP-triangulated points for initialization, following the CityGS evaluation protocol. We adopt the official MatrixCity-SmallCity densification setting as the default, and use the same setting for datasets not evaluated in the original {\grendelgs} paper for consistency.
Appendix~\ref{appendix:implementation_details} provides implementation details. We restrict comparison to methods sharing the vanilla 3DGS primitive, since LoBE-GS contributes only at the scene-decomposition layer. Closely related systems discussed in Sec. 2.3, CityGaussianV2, Momentum-GS, and CityGS-X, instead replace the primitive (2DGS, Scaffold-GS, Octree-GS), which would confound decomposition and primitive effects; Appendix~\ref{appendix:scope_of_baselines} details their relation to our method.

\input{figure_tex/qualitative_main}

\noindent \textbf{Evaluation Metrics and Runtime Protocol.} We evaluate reconstruction quality using PSNR, SSIM, and LPIPS, all computed without post-hoc color correction. Efficiency is measured by end-to-end wall-clock training time under a fixed GPU budget $K$, denoted as $T_{\mathrm{E2E}}^{(K)}$, using at most $K$ GPUs. This metric represents the total elapsed time required to produce the final reconstructed model, regardless of whether the method uses block-wise training or distributed training. For distributed baselines, $T_{\mathrm{E2E}}^{(K)}$ is simply the wall-clock time of the complete training run. For block-wise methods,
$T_{\mathrm{E2E}}^{(K)} = T_{\mathrm{coarse}} + T_{\mathrm{partition}} + T_{\mathrm{fine}}^{\mathrm{sched}}(K)$,
where $T_{\mathrm{fine}}^{\mathrm{sched}}(K)$ is the actual scheduled fine-stage runtime under the $K$-GPU budget. When the number of blocks $M$ exceeds $K$, blocks are queued in increasing block-ID order and dispatched to the first available GPU; therefore, $T_{\mathrm{fine}}^{\mathrm{sched}}(K)$ includes the full multi-wave execution time. We also report $\max_b T_{\mathrm{fine}}^{(b)}$ as a diagnostic load-balancing metric, capturing the slowest
individual block and potential stragglers.
For coarse-stage training, LoBE-GS uses 63 coarse blocks trained in parallel on MatrixCity-BigCity, while all other scenes use a single coarse model.
For block-wise runtime analysis, we follow the fine-stage block settings:
132 blocks for \emph{MatrixCity-BigCity},
36 blocks for \emph{MatrixCity-SmallCity}, 20 for \emph{Building}, 20 for \emph{Residence}, 9 for \emph{Rubble}, and 9 for \emph{Sci-Art}.

\noindent \textbf{System Configuration.} All experiments are conducted on a single compute node equipped with 8 NVIDIA L40 GPUs and 128 logical CPU cores from Intel Xeon Platinum 8362 CPUs. The actual GPU budget used by each method is reported with the corresponding result. Since all experiments are run within the same 8-GPU node, methods using the same GPU budget are evaluated under the same hardware setting.
\input{table_tex/runtime_breakdown}
\input{table_tex/component_ablation}
\subsection{Quantitative and Qualitative Results}
\label{sec:experiments:quantitative}
Table~\ref{tbl:quantitative_comparison} compares reconstruction quality and end-to-end runtime $T_{\text{E2E}}$ under an 8-GPU budget. Compared with 3DGS$^\dagger$, {\modelAbbrev} consistently improves PSNR, SSIM, and LPIPS across all evaluated scenes. Among partition-based methods, {\modelAbbrev} achieves the best overall quality--efficiency trade-off: it obtains competitive rendering quality, while achieving up to 9.1$\times$ speedup over {\citygs} and up to 3.7$\times$ speedup over {\vastgs}. On \textit{MatrixCity-BigCity}, {\gsmod} and
Grendel-GS achieve PSNR values of 22.64 and 23.01~dB, respectively, under the evaluated configurations.

Figure~\ref{fig:runtime_quality_tradeoff} further visualizes the runtime--quality trade-off under matched GPU budgets, showing that {\modelAbbrev} maintains competitive quality while achieving favorable wall-clock efficiency across 2, 4, and 8-GPU settings. As shown in Figure~\ref{fig:main_quality}, {\modelAbbrev} produces visual quality competitive with other methods under the same evaluation setting.
Additional comparisons of peak GPU memory consumption and final Gaussian counts across all methods are provided in Appendix~Table~\ref{tbl:quantitative_full}.

\subsection{Runtime Breakdown}
\label{sec:experiments:load_balance_and_runtime}

Table~\ref{tbl:runtime} reports the runtime breakdown of partition-based methods. We focus on $T_{\text{coarse}}$, $T_{\text{partition}}$, and $\max_b \Tfine[(b)]$, where $\Tfine[(b)]$ denotes the fine-stage optimization time of block $b$. The maximum fine-stage time $\max_b \Tfine[(b)]$ measures the slowest fine-training block and serves as a load-balancing metric. Since 3DGS$^\dagger$ and {\grendelgs} do not follow the same coarse-to-fine block-wise decomposition, their end-to-end runtimes are reported separately in Table~\ref{tbl:quantitative_comparison}.

{\modelAbbrev} consistently achieves the lowest $T_{\text{partition}}$, and $\max_b \Tfine[(b)]$ across all evaluated scenes. The reduction in $T_{\text{partition}}$ is especially pronounced, driven by fast camera selection, which redistributes only the parent's cameras rather than re-evaluating the full set per candidate block.
The reduction in $\max_b \Tfine[(b)]$ further shows that {\modelAbbrev} effectively suppresses fine-stage stragglers. This improvement comes from two complementary designs. First, the proposed load-balanced KD-tree partition with cutline refinement produces blocks with more balanced camera workloads. Second, visibility cropping and selective densification reduce the number of Gaussians involved in each block-wise optimization. Together, these components shorten the slowest fine-training block, which can introduce stragglers during parallel block-wise training.
End-to-end speedups reflect the complete pipelines rather than individual components. In particular, $T_{\text{coarse}}$ is not directly comparable: LoBE-GS uses Sparse Adam and trains 63 coarse blocks in parallel on MatrixCity-BigCity, whereas CityGS trains a single global model. To isolate partitioning gains, Appendix~\ref{appendix:partition_strategy} evaluates methods with a shared coarse checkpoint and identical visibility cropping and selective densification, reducing the maximum fine-stage runtime from 45 to 27 minutes on MatrixCity-SmallCity and from 40 to 28 minutes on Building relative to CityGS. Appendix~\ref{appendix:sensitivity} further evaluates coarse-model quality.

\input{figure_tex/proxy_correlation}

\subsection{Ablation Studies}
\label{sec:experiments:ablation}

To better understand the contribution of each design choice in {\modelAbbrev}, we conduct a series of ablation studies.
We first analyze the effect of the major components in our framework.
We then further investigate the proxy design used in load-balanced KD-tree partitioning, from both predictor quality and its impact on worst-block runtime and reconstruction quality.

\noindent \textbf{Evaluation of Components.} To quantify the contribution of each component, we conduct ablations on three representative datasets: MatrixCity-SmallCity, Residence, and Building.
Table~\ref{tbl:component_ablation} reports the maximum fine-stage runtime across blocks ($\max T_{\text{fine}}$), partition time ($T_{\text{partition}}$), peak GPU memory across blocks during training ($M_{\text{peak}}$), and reconstruction quality (PSNR/SSIM).
Besides ablating visibility cropping (\textbf{VC}), selective densification (\textbf{SD}), and cutline refinement (\textbf{CR}), we also include a \textit{w/o FCS} variant, where {\fcs} is replaced by the SSIM-based camera selection used in CityGS.

{\modelAbbrev} achieves the lowest $\max T_{\text{fine}}$ across all datasets while maintaining competitive memory usage and reconstruction quality.
Removing \textbf{VC} consistently increases both $\max T_{\text{fine}}$ and $M_{\text{peak}}$, showing that pruning invisible Gaussians is important for reducing optimization cost and memory usage.
Removing \textbf{SD} increases $\max T_{\text{fine}}$, indicating that unrestricted densification enlarges the active Gaussian set and amplifies computation in heavy blocks.
Replacing FCS with SSIM-based camera selection primarily increases $T_{\text{partition}}$, while also moderately increasing $\max T_{\text{fine}}$, due to differences in the selected camera sets.
Removing \textbf{CR} increases $\max T_{\text{fine}}$ on all datasets, confirming that cutline refinement further suppresses the fine-stage straggler beyond the initial KD-tree partition.
Across all variants, PSNR and SSIM vary only slightly, suggesting that the gains mainly come from better load control and reduced redundant computation rather than reduced reconstruction quality.

\noindent \textbf{Evaluation of Proxy.}
\label{sec:experiments:proxy_analysis} A key goal of load-balanced partitioning is to reduce the fine-stage \emph{straggler} effect, where training time is dominated by the slowest block. We therefore evaluate whether different proxies can identify heavy blocks and reduce the downstream worst-block runtime after KD-tree cutline refinement. We compare three single-proxy choices, \textbf{block area}, \textbf{visible points}, and \textbf{camera count}, along with several mixed-proxy variants.

We assess proxy quality using top-K\% heavy-block overlap (Overlap@K), defined as the fraction of the true top-K\% slowest blocks identified among the predicted top-K\% blocks ranked by each proxy. We also report the Pearson correlation $r^{(K)}$ between each proxy and measured fine-stage runtime within the true top-K\% slowest blocks. As shown in Table~\ref{tbl:topk-correlation} and Figure~\ref{fig:proxy_correlation}, camera count consistently achieves the highest overlap and strongest correlation across the evaluated thresholds, demonstrating its effectiveness in identifying potential stragglers.

\input{figure_tex/proxy_runtime}

We further evaluate whether this predictor-level advantage improves downstream partitioning. Figure~\ref{fig:proxy_runtime} compares random, single-proxy and mixed-proxy variants while keeping fast camera selection, visibility cropping, and selective densification enabled. For mixed proxies, each signal is normalized by its value at the KD-tree root node and summed as the cutline score. \textbf{Camera count} achieves the lowest worst-block runtime while preserving competitive reconstruction quality, motivating its use as our default proxy.

%% file: figure_tex/qualitative_main.tex
\begin{figure*}[!htbp]
\centering
\includegraphics[width=1.0\textwidth]{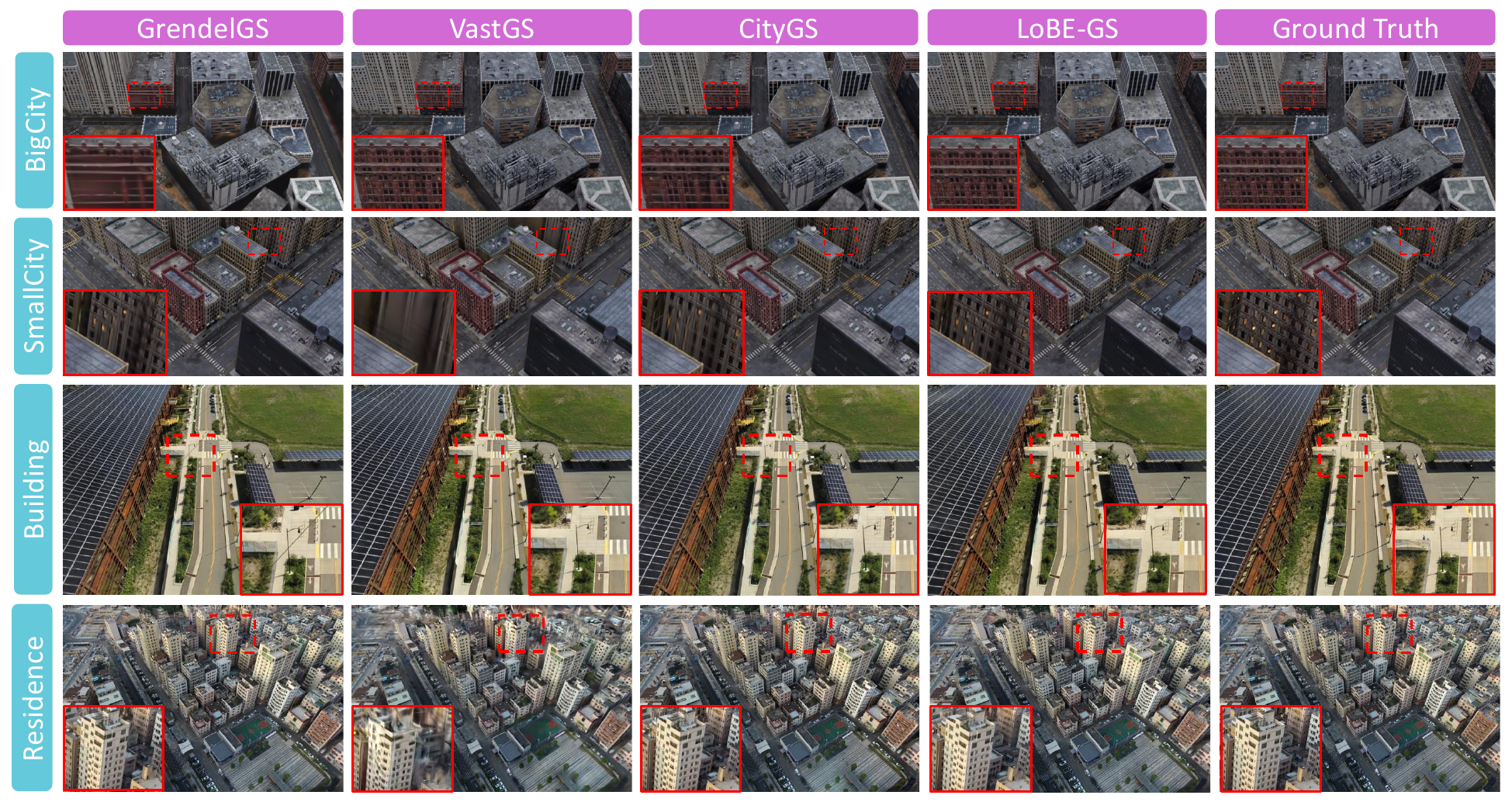}
\vspace{-7mm}
\caption{\textbf{Qualitative comparison on MatrixCity, Building and Residence datasets.} Full results are in Appendix~Fig.~\ref{fig:qualitative_full}}
\label{fig:main_quality}
\end{figure*}

%% file: table_tex/runtime_breakdown.tex
\begin{table*}[t]
\centering
\renewcommand{\arraystretch}{1.08}
\setlength{\tabcolsep}{4pt}
\resizebox{\textwidth}{!}{%
\begin{tabular}{l|ccc|ccc|ccc|ccc|ccc}
\toprule
\multirow{2}{*}{Method} &
\multicolumn{3}{c|}{MatrixCity-BigCity} &
\multicolumn{3}{c|}{MatrixCity-SmallCity} &
\multicolumn{3}{c|}{Building} &
\multicolumn{3}{c|}{Rubble} &
\multicolumn{3}{c}{Residence}\\
& $T_{\text{coarse}}$ & $T_{\text{partition}}$ & Max $T_{\text{fine}}$
& $T_{\text{coarse}}$ & $T_{\text{partition}}$ & Max $T_{\text{fine}}$
& $T_{\text{coarse}}$ & $T_{\text{partition}}$ & Max $T_{\text{fine}}$
& $T_{\text{coarse}}$ & $T_{\text{partition}}$ & Max $T_{\text{fine}}$
& $T_{\text{coarse}}$ & $T_{\text{partition}}$ & Max $T_{\text{fine}}$ \\
\midrule
VastGS
& -- & \underline{01:57} & 03:39
& -- & \underline{00:48} & 01:13
& -- & \underline{00:05} & \underline{00:44}
& -- & \underline{00:04} & \underline{00:39}
& -- & \underline{00:08} & \underline{00:49} \\
CityGS
& \underline{07:30} & 99:12 & \underline{02:41}
& \underline{00:52} & 01:39 & \underline{01:00}
& \underline{00:59} & 00:21 & 01:06
& \underline{01:06} & 00:09 & 01:14
& \underline{00:43} & 00:31 & 01:22 \\
{\modelAbbrev}
& \graybold{03:52} & \graybold{00:54} & \graybold{00:50}
& \graybold{00:38} & \graybold{00:08} & \graybold{00:27}
& \graybold{00:25} & \graybold{00:02} & \graybold{00:28}
& \graybold{00:23} & \graybold{00:01} & \graybold{00:25}
& \graybold{00:26} & \graybold{00:03} & \graybold{00:26} \\
\bottomrule
\end{tabular}%
}
\caption{\textbf{Runtime breakdown and load-balancing analysis for partition-based methods.}
For each dataset, we report coarse-stage optimization time $T_{\text{coarse}}$, partitioning time $T_{\text{partition}}$, and maximum fine-stage optimization time Max $T_{\text{fine}}$.
Max $T_{\text{fine}}$ measures the slowest fine-training block and serves as a load-balancing metric. A value of ``--'' indicates that the method does not include the corresponding stage.
All units are $hh$:$mm$.}
\label{tbl:runtime}
\end{table*}

%% file: table_tex/component_ablation.tex
\begin{table*}[t]
\centering
\small
\renewcommand{\arraystretch}{1.1}
\setlength{\tabcolsep}{3pt}
\resizebox{\linewidth}{!}{%
\begin{tabular}{l|ccccc|ccccc|ccccc}
\toprule
\multirow{2}{*}{Method} &
\multicolumn{5}{c|}{MatrixCity\text{-}SmallCity} &
\multicolumn{5}{c|}{Building} &
\multicolumn{5}{c}{Residence} \\
&
Max $T_{\text{fine}}$ & $T_{\text{partition}}$ & $M_{\text{peak}}$ & PSNR & SSIM &
Max $T_{\text{fine}}$ & $T_{\text{partition}}$ & $M_{\text{peak}}$ & PSNR & SSIM &
Max $T_{\text{fine}}$ & $T_{\text{partition}}$ & $M_{\text{peak}}$ & PSNR & SSIM\\
\midrule
w/o VC
& 01:39 & 00:08 & 22.1 & 27.74 & 0.876
& 01:00 & 00:02 & 13.4  & 21.76 & 0.785
& 00:51 & 00:03 & 11.9 & 21.35 & 0.800\\
w/o SD
& 00:41 & 00:08 & 9.2 & 27.72 & 0.875
& 00:46 & 00:02 & 10.6 & 21.63 & 0.781
& 00:44 & 00:03 & 9.4 & 21.45 & 0.803\\
w/o FCS
& 00:38 & 15:30 & 11.4 & 27.63 & 0.873
& 00:36 & 01:57 & 9.4  & 21.76 & 0.780
& 00:36 & 02:35 & 9.8 & 21.62 & 0.806\\
w/o CR
& 00:33 & 00:07 & 7.4 & 27.77 & 0.876
& 00:35 & 00:02 & 9.3 & 21.95 & 0.788
& 00:32 & 00:02 & \textbf{8.9} & 21.68 & 0.807 \\
\midrule
{\modelAbbrev}
& \textbf{00:27} & 00:08 & \textbf{7.4} & 27.69 & 0.875
& \textbf{00:28} & 00:02 & \textbf{9.3} & 22.13 & 0.790
& \textbf{00:26} & 00:03 & 9.0 & 21.48 & 0.805 \\
\bottomrule
\end{tabular}}
\caption{\textbf{Ablation on model components.}
We report the maximum fine-stage runtime across blocks ($\max T_{\text{fine}}$, $hh$:$mm$), partition time ($T_{\text{partition}}$, $hh$:$mm$), peak GPU memory across blocks ($M_{\text{peak}}$, GB), and rendering quality for {\modelAbbrev} and its ablated variants without visibility cropping (VC), selective densification (SD), fast camera selection (FCS), and cutline refinement (CR) in KD-tree partition.}
\label{tbl:component_ablation}
\end{table*}

%% file: figure_tex/proxy_correlation.tex
\begin{figure*}[t]
    \centering
    \includegraphics[width=0.9\linewidth]{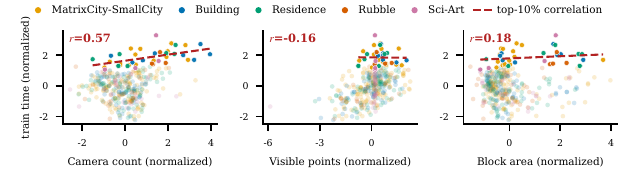}
    \vspace{-1em}
    \caption{
\textbf{Proxy correlation for the top-10\% heaviest blocks.}
Proxies and training time are $z$-normalized per scene before pooling. Colors denote datasets; highlighted points mark the heaviest blocks, with the dashed red line fitting this subset and $r$ denoting its Pearson correlation. Camera count correlates most strongly with training time.
}
    \label{fig:proxy_correlation}
\end{figure*}

%% file: figure_tex/proxy_runtime.tex
\begin{figure}[t]
    \centering
    \includegraphics[width=0.85\linewidth]{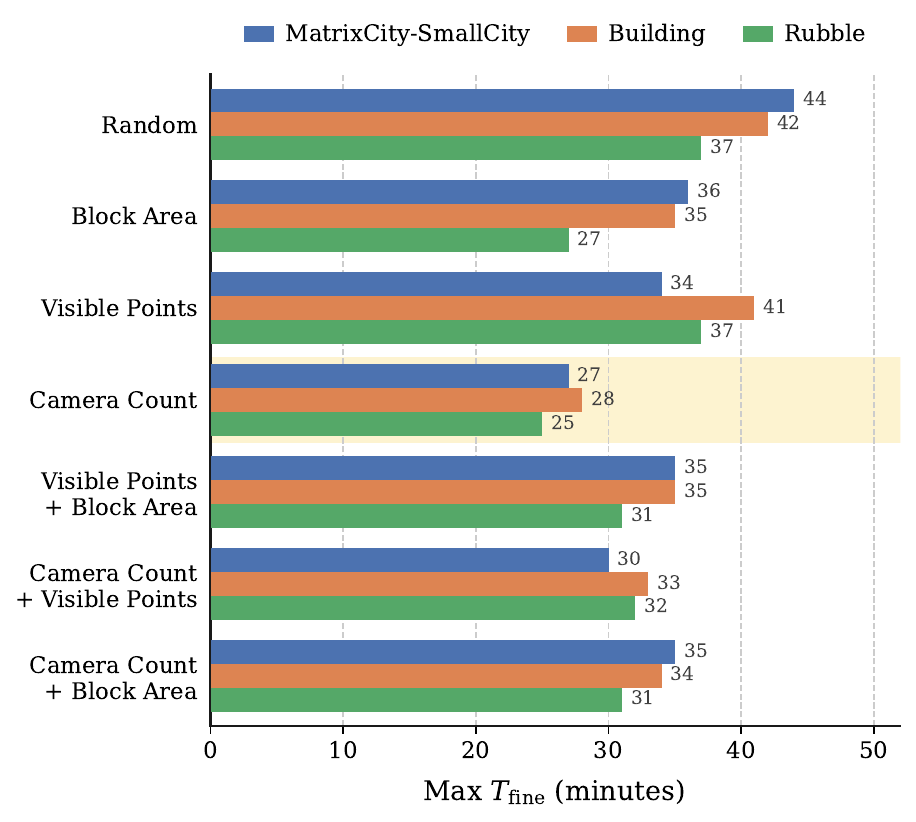}
    \vspace{-1em}
    \caption{
    \textbf{Worst-block fine-stage runtime across proxies.}
Max $T_{\text{fine}}$ under different KD-tree proxies, with FCS, VC, and SD enabled. Camera Count (highlighted) is lowest across all datasets.
}
    \label{fig:proxy_runtime}
\end{figure}

%% file: 5_Conclusion.tex
\section{Conclusion}
\label{sec:conclusion}
\input{table_tex/proxy_overlap}

We present {\modelAbbrev}, a load-balanced and efficient framework for parallel 3DGS training in large-scale scenes. By combining load-balanced KD-tree partitioning with fast camera selection, visibility cropping and selective densification, it reduces partition time and worst-block fine-stage runtime, achieving up to 9.1× faster end-to-end training than representative partition-based baselines at competitive reconstruction quality.

\noindent\textbf{Limitations.} Our approach has two main limitations.
First, the overall speedup still depends on the available GPU parallelism: when the number of GPUs is much smaller than the number of blocks, fine-training must be executed in multiple waves, which can limit the end-to-end gain.
Second, our camera-count proxy is primarily validated on aerial
scenes; its generalization to mixed-altitude
trajectories remains unexplored.

\noindent\textbf{Future work.} Since our block-wise load-balancing strategy is orthogonal to distributed 3DGS optimization systems, combining both directions is a promising path toward further acceleration.
Future work includes scaling to larger and more complex scenes, integrating LoD or 2DGS representations, extending the KD-tree partitioning with $z$-axis splits to better handle scenes with vertical extent.

%% file: table_tex/proxy_overlap.tex
\begin{table}[t]
    \centering
    \setlength{\tabcolsep}{6pt}
    \footnotesize
    \begin{tabular}{@{}l rrr rrr@{}}
        \toprule
        & \multicolumn{3}{c}{Overlap@$K$}
        & \multicolumn{3}{c}{$r^{(K)}$} \\
        \cmidrule(lr){2-4} \cmidrule(lr){5-7}
        $K$ & 5 & 10 & 20 & 10 & 20 & 30 \\
        \midrule
        Block area     & 0.35 & 0.41 & 0.31 & 0.18 & 0.43 & 0.41 \\
        Visible pts.   & 0.12 & 0.26 & 0.41 & $-0.16$ & 0.01 & 0.03 \\
        Camera count   & \textbf{0.53} & \textbf{0.47} & \textbf{0.44} & \textbf{0.57} & \textbf{0.51} & \textbf{0.52} \\
        \bottomrule
    \end{tabular}
    \caption{Heavy-block prediction quality measured by Overlap@K and Pearson correlation ($r^{(K)}$).}
    \label{tbl:topk-correlation}
\end{table}

%% file: 6_Acknowledgements.tex
\section{Acknowledgements}
We thank the anonymous reviewers for their valuable feedback. This work was supported by Taiwan's National
Science and Technology Council grants 114-2221-E-007-114-MY3,
113-2221-E-007-102-MY3, 114-2221-E-007-115-MY3, and 115-2634-F-007-003.
We also thank NVIDIA Corporation and NVIDIA AI Technology Center (NVAITC) for
providing access to the Taipei-1 supercomputer.

%% file: 7_Appendix.tex
This appendix provides details and analyses that support the main paper.
We first present implementation details in Sec.~\ref{appendix:implementation_details}. We then provide
additional quantitative and qualitative results in Sec.~\ref{appendix:additional_quantitative_results},
a comparison of partition strategies in Sec.~\ref{appendix:partition_strategy}, an
evaluation of our fast camera selection in Sec.~\ref{appendix:camera_selection}, an
analysis of sensitivity to coarse 3DGS quality in Sec.~\ref{appendix:sensitivity},
and load balance across datasets in Sec.~\ref{appendix:load_balance}.
Sec.~\ref{appendix:pose_appearance} studies camera pose re-estimation and appearance
modeling, Sec.~\ref{appendix:scope_of_baselines} details the scope of baselines, Sec.~\ref{appendix:interconnect} discusses interconnect sensitivity and deployment regime and Sec.~\ref{appendix:load_balance_kdtree_partition} details the load-balanced KD-tree
partitioning algorithm.

\section{Implementation Details}
\label{appendix:implementation_details}
\noindent \textbf{3DGS Training.} The coarse-training stage uses the Sparse Adam optimizer to accelerate training, which has minimal impact on final reconstruction quality. In contrast, the fine-training stage uses the standard Adam optimizer, as we found Sparse Adam to degrade performance in this setting. Except for \emph{selective densification}, the fine-training procedure follows the standard vanilla 3DGS pipeline, as adopted in {\citygs}. Our implementation is built on \emph{gsplat} (v1.5.3)~\cite{ye2025gsplat}, and uses \emph{fused-ssim}~\cite{mallick2024taming} for SSIM loss evaluation.

\noindent \textbf{Baseline Reproduction and Fairness.}
We take additional care to ensure that reproduced baselines are evaluated under fair and transparent settings. Since the official implementation of {\vastgs} is unavailable, we report results based on an unofficial implementation from \url{https://github.com/kangpeilun/VastGaussian}. However, we observe that directly applying this implementation to MatrixCity leads to substantially degraded reconstruction quality. Therefore, before partitioning, we apply an additional point cloud filtering step that removes sparse points with track length smaller than 3 or reprojection error larger than 5. This improves the quality of the partition input while keeping the training protocol unchanged. To make this modification transparent, we report both the original unofficial reproduction and our filtered version, denoted as \vastgs{} (unofficial) and \vastgs{} (ours), respectively, together with {\modelAbbrev} in Table~\ref{tbl:vastgs_reproduction}.

For {\grendelgs}, the original MatrixCity-SmallCity experiment uses dense depth-sampled points for initialization, which are available because MatrixCity-SmallCity is a synthetic dataset with ground-truth depth. In contrast, the CityGS evaluation protocol uses triangulated points, which is more consistent with real-world datasets where dense ground-truth depth is unavailable. Therefore, we use triangulated initialization as the main \grendelgs{} baseline in the main comparison, and additionally report the depth-initialized variant as \grendelgs{}$^\dagger$ in Table~\ref{tbl:grendelgs_initialization}. Specifically, \grendelgs{} uses 3.8M triangulated points, while \grendelgs{}$^\dagger$ uses the original 15.5M depth-sampled initialization. Following the configuration demonstrated in the official repository, we set the batch size equal to the GPU count (i.e., 8 for our 8-GPU experiments). Both variants use the same author-provided official densification setting: 400K training iterations, densification from 20K to 150K iterations with a 1K interval, and opacity reset every 30K iterations. On BigCity we double Grendel-GS's official schedule to 800K iterations with densification to 400K. This comparison separates the effect of the reconstruction method from the effect of stronger synthetic-data initialization.

\input{table_tex/reproduce_details}
\input{table_tex/quantitative_full}
\noindent \textbf{MatrixCity-BigCity Setup.}
The scene spans roughly $25~\text{km}^2$ with about $50{,}000$ images
($\sim\!10\times$ MatrixCity-SmallCity), well beyond the extent and image counts
at which partition-based methods (e.g., \vastgs, \citygs) and distributed baseline (e.g. \grendelgs) are typically evaluated. At this scale a single coarse model is no longer computationally
efficient, so we partition \emph{both} stages --- the $N_{\mathrm{coarse}}>1$
case of Sec.~\ref{sec:methodology:load-balance-aware_scene_partition}. Since FCS needs depth from an existing coarse model, we first
split the scene into coarse blocks via \vastgs's camera selection with our
load-balanced KD-tree partition and train them in parallel for $30$K iterations,
then re-partition more finely with FCS under the same scheme and train the fine
models for another $30$K iterations. This is not fundamental: the framework only
requires \emph{some} initial coarse reconstruction. As the KD-tree terminates on
a per-node camera budget, we cap cameras per block ($<\!2{,}000$ coarse,
$<\!1{,}000$ fine) to control per-GPU memory and compute, yielding $63$ coarse
and $132$ fine blocks. The scale ceiling of {\modelAbbrev} is thus set by the
per-block camera budget, not by any single model's capacity.

\noindent \textbf{Reproducibility.} All reported runtimes are measured within the Docker environment to reduce discrepancies caused by library mismatches or system-level variations.

\noindent \textbf{Declaration of LLM Usage.} Large Language Models (LLMs) are used only for grammar editing.

\section{More Quantitative and Qualitative Results}
\label{appendix:additional_quantitative_results}
\noindent\textbf{More Quantitative Results.}
Table~\ref{tbl:quantitative_full} reports full results, including Gaussian counts
(\#GS) and $M_{\mathrm{peak}}$, the peak GPU memory across all blocks.

The baselines trace a clear capacity--memory trade-off: \citygs{} incurs the
largest $M_{\mathrm{peak}}$ among partition-based methods on most scenes, while
\vastgs{} keeps $M_{\mathrm{peak}}$ low only by producing substantially fewer
Gaussians. {\modelAbbrev} breaks this trade-off, matching or exceeding \citygs's
Gaussian count while using roughly half the peak memory on the MatrixCity scenes.
This indicates that our efficiency gains come from a smaller per-block working
set rather than from sacrificing model capacity.

Figure~\ref{fig:appendix_runtime_quality_tradeoff} plots the quality--runtime trade-off
over five datasets, four methods, and three GPU budgets, each curve connecting the
2-, 4-, and 8-GPU points of one method. \grendelgs{} attains strong PSNR on
Building, Residence, and Sci-Art but weaker SSIM and LPIPS, so high PSNR does not
imply better structural or perceptual quality; \citygs{} is strong on SSIM and
LPIPS but slow under small GPU budgets, while \vastgs{} is faster yet
inconsistent. {\modelAbbrev} is competitive on all three quality metrics while consistently running faster than prior partition-based methods at the same GPU budget, owing to its reduced partitioning and fine-training overhead.
\input{table_tex/runtime_quality_full}

\noindent\textbf{More Qualitative Results.}
Figure~\ref{fig:qualitative_full} shows novel-view renderings for all methods and
datasets. {\modelAbbrev} remains visually competitive with the baselines across
scenes, and on MatrixCity-BigCity it produces noticeably sharper reconstructions
than \grendelgs{}.
\input{figure_tex/qualitative_full}

\noindent\textbf{Block Boundary Evaluation.}
Since blocks are trained independently, a viewpoint spanning several of them could
reveal seams. As shown in Figure~\ref{fig:block_boundary}, the rendered views exhibit
no obvious discontinuities at block boundaries in the evaluated examples.

\input{figure_tex/qualitative_block_boundary}

\section{Evaluation of Different Partition Strategies}
\label{appendix:partition_strategy}
To isolate the effect of the partition strategy, we \textbf{fix the coarse model} and optimization setup across all methods: all variants are trained from the same coarse model and use the same visibility cropping and selective densification (VC+SD) configuration. We only vary the data partition strategy between {\vastgs}, {\citygs}, and our load-balanced KD-tree partitioning.
As summarized in Table~\ref{tbl:partition_strategies}, our method consistently achieves the lowest maximum fine-stage time (Max $T_{\text{fine}}$) on all datasets, while maintaining comparable or better reconstruction quality. These results indicate that, under a shared coarse model and VC+SD setup, our partition strategy offers a more favorable quality versus Max-$T_{\text{fine}}$ trade-off.
\input{table_tex/partition_strategies}

\section{Evaluation of Fast Camera Selection}
\label{appendix:camera_selection}

\noindent\textbf{Agreement with SSIM-based selection.}
The three camera selection strategies differ in how visibility is estimated.
\citygs{} measures each camera's SSIM contribution to a block, which requires a
separate render pass per block; \vastgs{} projects the eight corners of a block's
bounding box into the image and tests coverage against their convex hull; FCS
resolves all blocks from a single depth render. To verify that our approximation
preserves the selection itself, we treat the \citygs{} SSIM-based selection as a
reference and report precision, recall, and F1 for FCS and the \vastgs{}
bounding-box heuristic on a fixed coarse checkpoint
reported in Table~\ref{tbl:camera_agreement}.

FCS attains the higher F1 on four of five scenes, with recall above $0.85$
everywhere, indicating that it rarely discards a camera the reference would have
kept. Its precision is lower on MatrixCity-SmallCity ($0.641$), meaning FCS admits
some extra cameras --- the conservative failure direction, since surplus
supporting views cost training time but do not remove supervision. The
bounding-box heuristic is competitive on Rubble but degrades sharply on
MatrixCity-SmallCity ($0.499$ F1), because the convex-hull test is not occlusion
aware: a camera is counted as supporting a block whenever the hull falls inside
the frame, regardless of whether the block is actually visible. Under the dense
vertical structure of MatrixCity, distant blocks are frequently occluded by
foreground buildings, so the criterion admits many cameras that contribute nothing
to the block, yielding low precision ($0.340$) at comparable recall.

\input{table_tex/cam_sel_aggreement}
\input{table_tex/camera_selection_sensitivity}
\noindent\textbf{Partitioning cost.}
Let $N_{\mathrm{cam}}$ and $N_{\mathrm{block}}$ denote the number of cameras and
blocks, and $C_{\mathrm{render}}$ the cost of one render pass. The three
strategies differ sharply in asymptotic cost. \citygs{} renders each camera
once per block, costing
$O(N_{\mathrm{cam}} \cdot N_{\mathrm{block}} \cdot C_{\mathrm{render}})$;
\vastgs{} avoids rendering entirely and only projects eight box corners per
camera--block pair, costing $O(N_{\mathrm{cam}} \cdot N_{\mathrm{block}})$; FCS
renders each camera once and resolves all blocks from that single depth map,
costing $O(N_{\mathrm{cam}} \cdot (C_{\mathrm{render}} + N_{\mathrm{block}}))$.
FCS therefore pays the render cost only once per camera rather than once per
camera--block pair, retaining occlusion awareness at a cost that is asymptotically
closer to the projection-only heuristic than to SSIM contribution testing.
Figure~\ref{fig:partition-time} matches this analysis: only \citygs{} grows with
block count in Figure~\ref{fig:partition-time}(a), \vastgs{} also grows linearly in block count, but its per-pair constant — eight corner projections and a convex-hull test — is several orders of magnitude below a render, so its curve is dominated by M-independent preprocessing and appears flat at this vertical scale. FCS stays
near-flat, and all three scale linearly in camera count with markedly different
constants. Note the differing vertical ranges between the two panels.

\noindent\textbf{Selection threshold $\tau$.}
Table~\ref{tbl:tau} varies $\tau$ under $9$/$16$/$36$ blocks on
MatrixCity-SmallCity. Quality is stable overall (PSNR spread $\leq 0.2$\,dB
within each block count), but the best setting shifts with granularity: coarser
partitions favor a larger $\tau$ ($0.20$ at $9$ blocks) while finer partitions
favor a smaller one ($0.05$ at $36$ blocks). This is expected, since smaller
blocks subtend a smaller fraction of each view, so a fixed ratio prunes supporting
cameras more aggressively as the block count grows. We therefore use
$\tau = 0.10$ as a default for the moderate block counts used in our main
experiments. Our results suggest that a larger threshold may be preferable for coarser partitions, whereas a smaller threshold may better preserve supporting views for finer partitions. Partition time is unaffected by $\tau$, as the cost is dominated
by scoring all camera--block pairs.
\input{figure_tex/partition_scalability}

\section{Impact of Coarse Model Quality}
\label{appendix:sensitivity}
{\modelAbbrev} uses a coarse 3DGS model to guide scene partitioning and camera
selection. To test how sensitive the framework is to this prior, we vary the number
of coarse training iterations and re-run the full pipeline on MatrixCity, Building,
and Rubble, deriving the partitioning and camera selection from the corresponding
coarse model while keeping all other components fixed.
Table~\ref{tbl:coarse_sensitivity} shows that a longer coarse budget yields a
markedly stronger coarse reconstruction with more Gaussians, yet the final
fine-stage metrics stay nearly unchanged across budgets. Even an under-trained
coarse prior therefore provides sufficiently accurate structural and visibility
cues for partitioning and camera selection, and improving it further brings
diminishing returns at the fine stage.
Overall, the results indicate that the partitioning and fine optimization stages of LoBE-GS are robust to moderate inaccuracies in the coarse prior, supporting the practicality of using a moderately trained coarse model in practice.

\input{table_tex/coarse_model_sensitivity}

\section{Load Balance Across Datasets}
\label{appendix:load_balance}
Figure~\ref{fig:loadbalance_vis} compares the per-block fine-stage runtime distributions of \citygs{}, {{\modelAbbrev}}  w/o cut refinement (CR), and the full  {{\modelAbbrev}}  across all evaluated datasets.
Both variants of {{\modelAbbrev}} produce a more balanced workload distribution than \citygs{}, with lower dispersion and a reduced runtime tail, indicating fewer straggler blocks during parallel fine training.
Notably, {{\modelAbbrev}} w/o CR already provides a clear improvement, showing that the proposed load balance-aware partitioning is effective on its own.

The full {{\modelAbbrev}} further tightens the runtime distribution and reduces the worst-case block runtime, demonstrating that CR offers an additional gain by better equalizing workloads across adjacent partitions.
Overall, {{\modelAbbrev}}  achieves the most uniform per-block runtime distribution among the compared methods.

These consistent trends across datasets demonstrate the robustness of our partitioning and refinement strategy.
By alleviating workload skew and reducing stragglers, our method improves GPU utilization and enhances end-to-end training efficiency.
\input{figure_tex/load_balance_vis}

\input{table_tex/cam_pose_bilateral_grid}

\section{Camera Pose Re-estimation and Appearance Modeling}
\label{appendix:pose_appearance}

The Mill-19 and UrbanScene3D captures provide camera parameters derived from
onboard drone GPS and IMU. MegaNeRF~\cite{Turki2022CVPR} reports these to be
insufficiently accurate for high-fidelity reconstruction---models trained on the
raw sensor poses were severely blurred---and identifies pose accuracy as the
single largest limiting factor for large-scale view synthesis, refining the
poses with PixSFM~\cite{lindenberger2021pixsfm} before training. 3D Gaussian
Splatting is particularly sensitive to such error: residual misalignment cannot
be reconciled across views and is instead absorbed as inflated,
view-inconsistent primitives. In this section, we therefore re-estimate all camera parameters
offline with COLMAP~\cite{schoenberger2016sfm} prior to training and hold them
fixed throughout optimization. As shown in
Table~\ref{tab:appendix_pose_appearance}, this yields consistent gains over the
released parameters, indicating that pose quality remains a bottleneck at this
scale. Please note that the experiments in the main paper follow the standard benchmark setting and use the released camera parameters without re-estimation.

\paragraph{Coarse-to-fine appearance optimization.}
To support appearance modeling, we follow
the bilateral-grid formulation of~\cite{wang2024bilateral} to attach a per-image bilateral grid to the rendering pipeline. Each training image $i$ is assigned a grid
$A^i \in \mathbb{R}^{W \times H \times M \times 12}$ whose cells store the
coefficients of a $3 \times 4$ affine color transform. Slicing $A^i$ at each
pixel, guided by the rendered color, maps the rendered image
$\mathcal{I}^r_i$ to its post-processed counterpart $\mathcal{I}^a_i$. The grids
are optimized jointly with the Gaussians and discarded at inference.

We integrate this into our coarse-to-fine pipeline in two stages. In the
\emph{coarse} stage, a global set of per-image grids $\{A^i\}_{i=1}^{L}$ is
optimized jointly with the coarse model over all $L$ training views, capturing
the scene-wide photometric drift of the capture session. In the \emph{fine}
stage, the scene is partitioned into blocks, and each block inherits the
corresponding global grids as initialization and fine-tunes them on its own
image subset. This allows each block to adapt to the local illumination
conditions of the region it covers while remaining anchored to a shared global
appearance, avoiding the color discontinuities that arise when independent
per-block grids are trained from scratch.

Following VastGaussian~\cite{lin2024vastgaussian}, we apply the photometric and
structural terms to different images: the $\mathcal{L}_1$ term is computed on the
post-processed image $\mathcal{I}^a_i$, allowing the appearance module to absorb
per-image photometric variation, while the $\mathcal{L}_{\text{D-SSIM}}$ term is computed on the raw
render $\mathcal{I}^r_i$, so that structural supervision acts directly on the
geometry rather than on the appearance-corrected output. Both stages therefore
minimize
\begin{equation}
\mathcal{L} = (1-\lambda)\,\mathcal{L}_1\bigl(\mathcal{I}^a_i, \mathcal{I}_i\bigr)
\;+\; \lambda\,\mathcal{L}_{\text{D-SSIM}}\bigl(\mathcal{I}^r_i, \mathcal{I}_i\bigr)
\;+\; \lambda_{\mathrm{TV}}\,\mathcal{L}_{\mathrm{TV}},
\label{eq:appearance_loss}
\end{equation}
where $\mathcal{I}_i$ is the ground-truth image and $\mathcal{L}_{\mathrm{TV}}$
is the total variation term of~\cite{wang2024bilateral} regularizing grid
smoothness.

The bilateral grid is a training-time module and is not applied at test time,
since the appearance parameters of a held-out view are unknown. We therefore
report color-corrected (CC) metrics, which fit a global color correction before
evaluation and form the meaningful basis of comparison here.

\input{figure_tex/qualitative_colmap_bilateral_grid}

\paragraph{Qualitative results.}
Figure~\ref{fig:qualitative} illustrates the contribution of each component.
Renders trained on the originally distributed poses exhibit pronounced blurring
on high-frequency structure, as reprojection error is absorbed into oversized,
view-inconsistent primitives. Re-estimating the poses with COLMAP largely
resolves this blurring, but the resulting reconstructions still exhibit floaters
and low-frequency color blotching: with the geometry now correctly aligned, the
remaining photometric inconsistency across training images must instead be
explained by spurious view-dependent structure. Combining both components
addresses each failure mode in turn---the re-estimated poses recover sharp
high-frequency detail, while the bilateral grid absorbs the per-image
photometric variation, suppressing floaters and color blotching.
\paragraph{Quantitative results.}
Table~\ref{tab:appendix_pose_appearance} follows the same trend, with one caveat
in how it must be read: the bilateral grid lowers raw PSNR in every scene while
improving LPIPS, and this gap closes under color correction. This is expected.
The grid absorbs photometric variation during training, so the scene converges
to a canonical appearance misaligned with the arbitrary exposure of any held-out
view. Raw PSNR, dominated by low-frequency intensity offsets, penalizes this
mismatch even when the reconstruction is structurally better, whereas LPIPS is
largely insensitive to global color shifts. The color-corrected metrics resolve
the ambiguity: Residence and SciArt improve on all three CC metrics (up to
$+1.42$~dB CC-PSNR on SciArt), while Building and Rubble show slightly lower
CC-PSNR ($-0.66$ and $-0.36$~dB) alongside better CC-LPIPS.

\section{Scope of Baselines}
\label{appendix:scope_of_baselines}
Large-scale 3DGS systems vary along two largely independent axes: the decomposition and scheduling layer, which determines how the scene is split, how views are assigned, and how submodels are scheduled; and the primitive layer, which determines what each submodel optimizes. LoBE-GS contributes only to the former and deliberately retains vanilla 3DGS as the primitive, so that measured gains are attributable to partitioning and scheduling alone. We therefore compare against methods sharing this primitive: CityGS and VastGS (partition-based), Grendel-GS (distributed), and $\text{3DGS}^{\dagger}$.

Several closely related systems in main paper Sec.~2.3 change the primitive instead: CityGaussianV2 adopts CityGS's pipeline with 2DGS primitives and a modified densification rule; Momentum-GS also adopts CityGS's pipeline with Scaffold-GS primitives, where Gaussians are MLP-decoded from anchors; and CityGS-X builds on Grendel-GS's distributed system with Octree-GS, a multi-level Scaffold-GS variant. Each primitive changes per-iteration cost, memory, densification dynamics, and achievable quality independently of how the scene is decomposed, so a head-to-head comparison would confound the two axes and could not isolate the partition strategy — the quantity this paper is about. We regard these methods as complementary: because LoBE-GS operates purely at the decomposition layer, our partitioning and camera selection are in principle substitutable into any pipeline that splits a scene and assigns cameras to blocks, including the CityGS-derived pipelines above.

\section{Interconnect Sensitivity and Deployment Regime}
\label{appendix:interconnect}
All experiments run on a single node of 8 NVIDIA L40 GPUs over PCIe, without
NVLink --- the regime our design targets. Distributed systems such as
Grendel-GS exchange Gaussian state every iteration, so on NVLink or DGX-class
hardware the advantage of a communication-free design narrows. We claim no
advantage there.

What does not depend on the interconnect is fault tolerance. Blocks are
independent processes: if a device fails or is preempted, only the block on it
is re-run and all completed blocks are retained. A synchronized run instead
loses the in-flight work of every device and must restart from the last
checkpoint with the full device set restored. This matters most for long runs
on shared or preemptible hardware --- MatrixCity-BigCity takes 132 fine blocks
over ${\sim}15$ hours --- where the probability of losing a device at some
point is not small. The same independence lets one partition be rescheduled
onto any number of GPUs. The coarse
stage remains a single point of failure, so only the fine stage, which
dominates runtime, is fault-isolated. We report this as a deployment property,
not a measured contribution, since we do not evaluate failure recovery.

\section{Algorithm of Load-balanced KD-tree scene partition}
\label{appendix:load_balance_kdtree_partition}
Algorithm~\ref{alg:kd_tree_partition} summarizes the full procedure of our load-balanced KD-tree scene partitioning.

\input{table_tex/algo}

%% file: table_tex/reproduce_details.tex
\begin{table}[t]
\centering
\scriptsize

\resizebox{\linewidth}{!}{%
\begin{tabular}{lcccc}
\toprule
Method & PSNR & SSIM & LPIPS & $T_{\text{E2E}}^{(8)}$ \\
\midrule
{\vastgs} (unofficial) & 24.64 & 0.761 & 0.351 & 5:32 \\
{\vastgs} (ours)       & 26.92 & 0.845 & 0.235 & 4:31 \\
{\modelAbbrev}         & 27.69 & 0.875 & 0.185 & 2:08 \\
\bottomrule
\end{tabular}}
\caption{\textbf{Effect of {\vastgs} reproduction on MatrixCity-SmallCity.}
Effect of our point filtering for the unofficial {\vastgs} reproduction.}
\label{tbl:vastgs_reproduction}
\end{table}

\begin{table}[t]
\centering
\scriptsize

\resizebox{\linewidth}{!}{%
\begin{tabular}{llccc}
\toprule
Method & Init. & PSNR & SSIM & LPIPS \\
\midrule
{\grendelgs}           & Triangulate & 26.96 & 0.850 & 0.224 \\
{\grendelgs}$^\dagger$ & Depth  & 27.32 & 0.862 & 0.197 \\
{\modelAbbrev}         & Triangulate & 27.69 & 0.875 & 0.185 \\
\bottomrule
\end{tabular}}
\caption{\textbf{Effect of {\grendelgs} initialization on MatrixCity-SmallCity.}
Effect of dense depth-sampled initialization for {\grendelgs}.}
\label{tbl:grendelgs_initialization}
\end{table}

%% file: table_tex/quantitative_full.tex
\begin{table*}[t!]
\centering
\renewcommand{\arraystretch}{1.1}
\setlength{\tabcolsep}{3pt}

\resizebox{\textwidth}{!}{%
\begin{tabular}{l|ccccrr|ccccrr|ccccrr}
\toprule
\multirow{2}{*}{Method} &
\multicolumn{6}{c|}{MatrixCity-BigCity} &
\multicolumn{6}{c|}{MatrixCity-SmallCity} &
\multicolumn{6}{c}{Building} \\
& PSNR & SSIM & LPIPS & $T_{\text{E2E}}^{(8)}$ & $M_{\text{peak}}$ & \#GS
& PSNR & SSIM & LPIPS & $T_{\text{E2E}}^{(8)}$ & $M_{\text{peak}}$ & \#GS
& PSNR & SSIM & LPIPS & $T_{\text{E2E}}^{(8)}$ & $M_{\text{peak}}$ & \#GS \\
\midrule
\multicolumn{19}{l}{\textbf{Single-model / distributed methods}} \\
{\gsmod}
& 22.64 & 0.660 & 0.454 & 24:29 & 16.53 & 10.5
& 25.57 & 0.794 & 0.315 & 02:10 & 17.06 & 11.3
& 20.46 & 0.720 & 0.305 & 01:30 & 9.94 & 6.2 \\
\grendelgs
& 23.01 & 0.740 & 0.354 & 10:46 & 21.86 & 68.1
& \third{26.96} & \third{0.850} & \third{0.224} & 02:05 & 10.99 & 20.8
& \best{22.52} & \third{0.764} & \third{0.277} & 00:46 & 8.18 & 6.9 \\
\midrule
\multicolumn{19}{l}{\textbf{Partition-based methods}} \\
\vastgs
& \best{29.17} & \best{0.889} & \second{0.173} & 57:29 & 20.10 & 196.5
& 26.92 & 0.845 & 0.235 & 04:31 & 11.71 & 19.1
& 21.31 & 0.743 & 0.277 & 02:28 & 6.73 & 10.5 \\
\citygs
& \third{27.87} & \third{0.838} & \third{0.265} & 140:31 & 17.20 & 58.4
& \second{27.46} & \second{0.865} & \second{0.204} & 06:53 & 14.11 & 23.6
& \third{21.55} & \second{0.778} & \second{0.246} & 04:51 & 20.39 & 13.2 \\
{\modelAbbrev}
& \second{28.88} & \second{0.888} & \best{0.170} & \graybold{15:27} & 11.50 & 197.6
& \best{27.69} & \best{0.875} & \best{0.185} & \graybold{02:08} & 7.42 & 28.9
& \second{22.13} & \best{0.790} & \best{0.231} & \graybold{01:33} & 9.34 & 13.9 \\
\bottomrule
\end{tabular}}

\vspace{2mm}

\resizebox{\textwidth}{!}{%
\begin{tabular}{l|ccccrr|ccccrr|ccccrr}
\toprule
\multirow{2}{*}{Method} &
\multicolumn{6}{c|}{Rubble} &
\multicolumn{6}{c|}{Residence} &
\multicolumn{6}{c}{Sci-Art} \\
& PSNR & SSIM & LPIPS & $T_{\text{E2E}}^{(8)}$ & $M_{\text{peak}}$ & \#GS
& PSNR & SSIM & LPIPS & $T_{\text{E2E}}^{(8)}$ & $M_{\text{peak}}$ & \#GS
& PSNR & SSIM & LPIPS & $T_{\text{E2E}}^{(8)}$ & $M_{\text{peak}}$ & \#GS \\
\midrule
\multicolumn{19}{l}{\textbf{Single-model / distributed methods}} \\
{\gsmod}
& 25.47 & \third{0.777} & \third{0.277} & 01:10 & 5.84 & 3.7
& 21.44 & 0.791 & 0.236 & 01:22 & 6.63 & 4.1
& 21.34 & 0.833 & 0.232 & 00:48 & 3.64 & 2.1 \\
\grendelgs
& \third{25.72} & 0.768 & 0.301 & 00:30 & 5.90 & 3.2
& \best{22.37} & \third{0.804} & \third{0.227} & 00:43 & 10.99 & 4.8
& \best{22.76} & \best{0.847} & \second{0.222} & 00:33 & 11.20 & 2.0 \\
\midrule
\multicolumn{19}{l}{\textbf{Partition-based methods}} \\
\vastgs
& 25.13 & 0.749 & 0.316 & 01:15 & 3.75 & 5.3
& 20.88 & 0.778 & 0.242 & 02:28 & 4.74 & 7.9
& 20.88 & 0.796 & 0.291 & 01:33 & 3.20 & 2.9 \\
\citygs
& \second{25.77} & \best{0.813} & \best{0.228} & 03:33 & 14.56 & 9.7
& \second{22.00} & \best{0.813} & \second{0.211} & 06:11 & 21.66 & 10.8
& \third{21.39} & \third{0.837} & \third{0.230} & 02:52 & 14.04 & 3.7 \\
{\modelAbbrev}
& \best{25.93} & \best{0.813} & \second{0.232} & \graybold{01:01} & 9.40 & 9.2
& \third{21.48} & \second{0.805} & \best{0.209} & \graybold{01:38} & 9.00 & 11.2
& \second{21.45} & \second{0.843} & \best{0.219} & \graybold{00:49} & 5.49 & 4.5 \\
\bottomrule
\end{tabular}}

\vspace{-2mm}
\caption{\textbf{Quantitative comparison on MatrixCity, Mill19, and UrbanScene3D datasets.}
We report PSNR, SSIM, LPIPS, end-to-end runtime $T_{\text{E2E}}^{(8)}$, peak
per-GPU memory $M_{\text{peak}}$ (GB), and final model size \#GS (millions of
Gaussians) under an 8-GPU budget.
The \bestlegend{best}, \secondlegend{second}, and \thirdlegend{third} best reconstruction quality are highlighted across five methods.
For runtime and memory, the best among partition-based methods is highlighted in \graylegend{bold}. All runtimes are reported in $hh$:$mm$.
}
\label{tbl:quantitative_full}
\end{table*}

%% file: table_tex/runtime_quality_full.tex
\begin{figure*}[!htbp]
\centering
\includegraphics[width=0.75\textwidth]{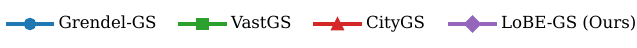}\\[1mm]
\setlength{\tabcolsep}{2pt}
\newcommand{\rqfig}[1]{\raisebox{-0.5\height}{\includegraphics[width=0.25\textwidth]{#1}}}
\newcommand{\rowname}[1]{\rotatebox[origin=c]{90}{#1}}
\begin{tabular}{@{}c ccc@{}}
 & \hspace{6mm}PSNR & \hspace{6mm}SSIM & \hspace{6mm}LPIPS \\[0.5mm]
\rowname{MatrixCity-SmallCity} &
\rqfig{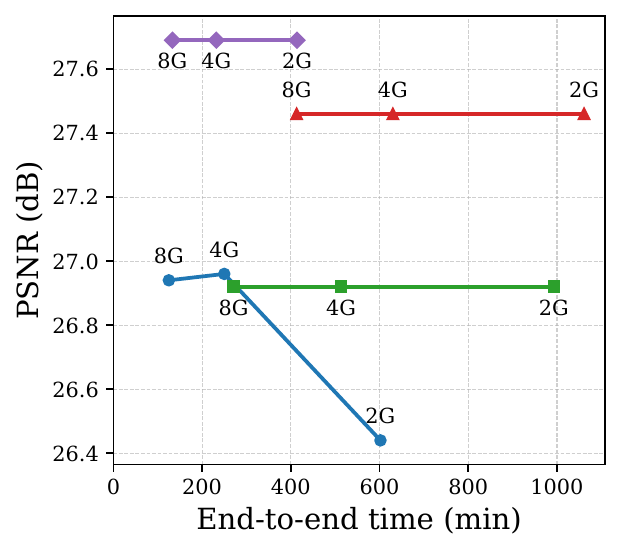} &
\rqfig{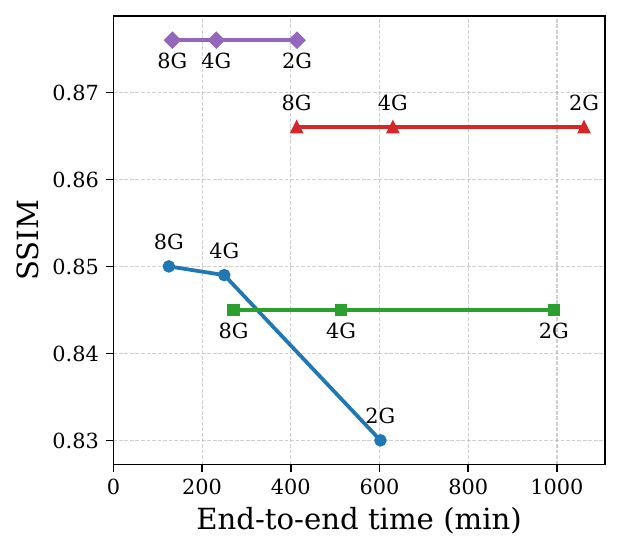} &
\rqfig{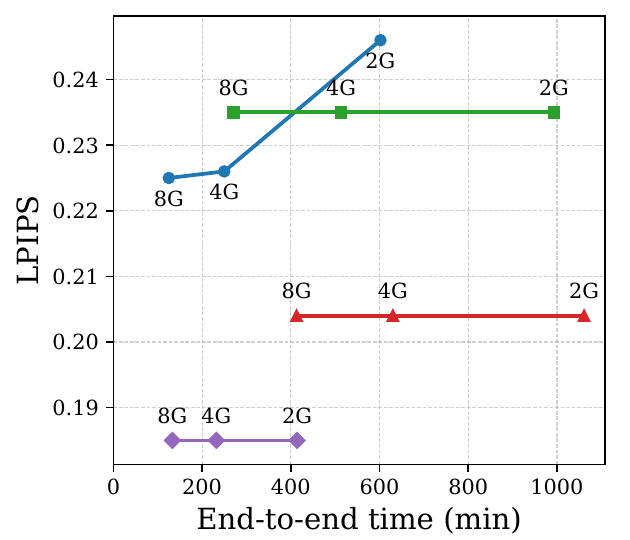} \\[2.5mm]
\rowname{Building} &
\rqfig{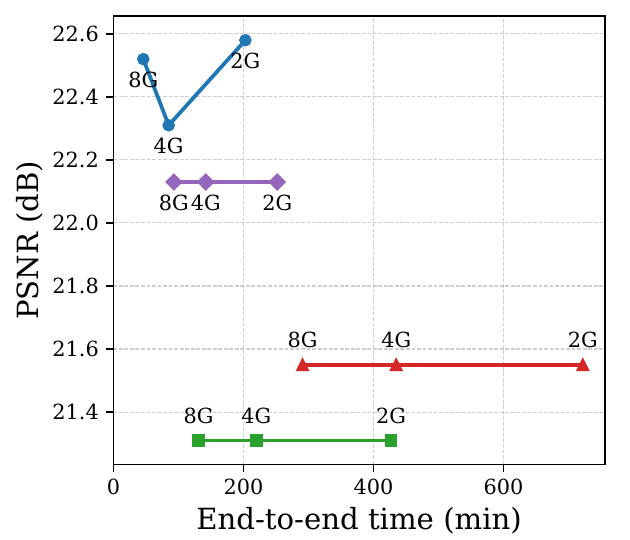} &
\rqfig{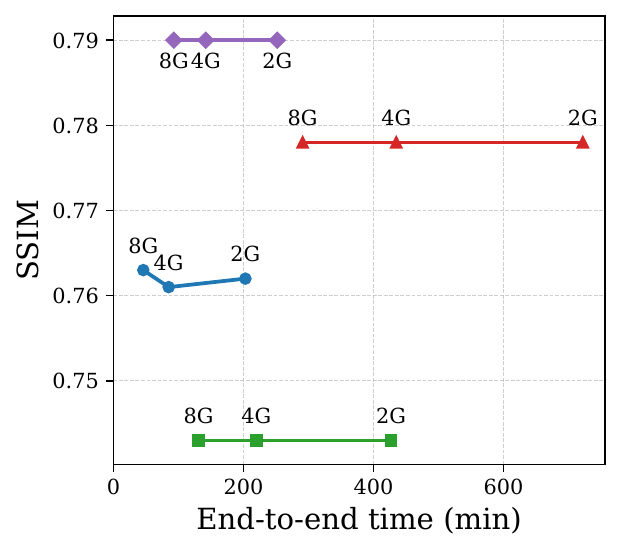} &
\rqfig{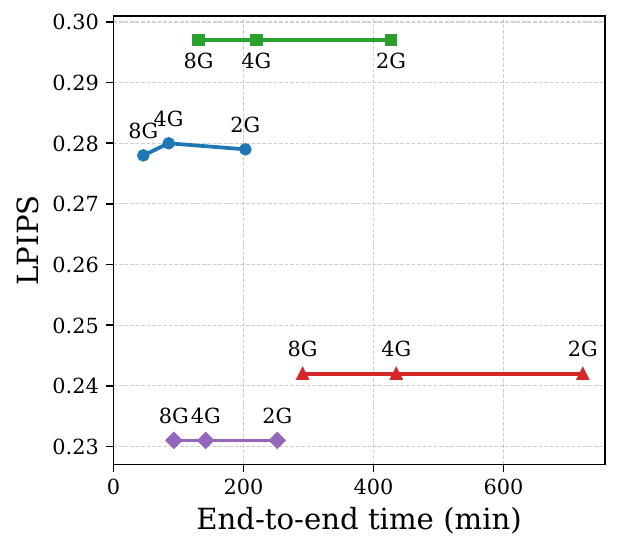} \\[2.5mm]
\rowname{Rubble} &
\rqfig{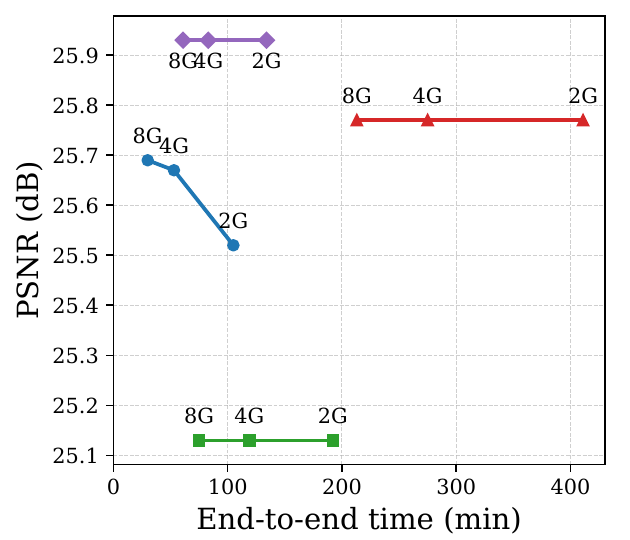} &
\rqfig{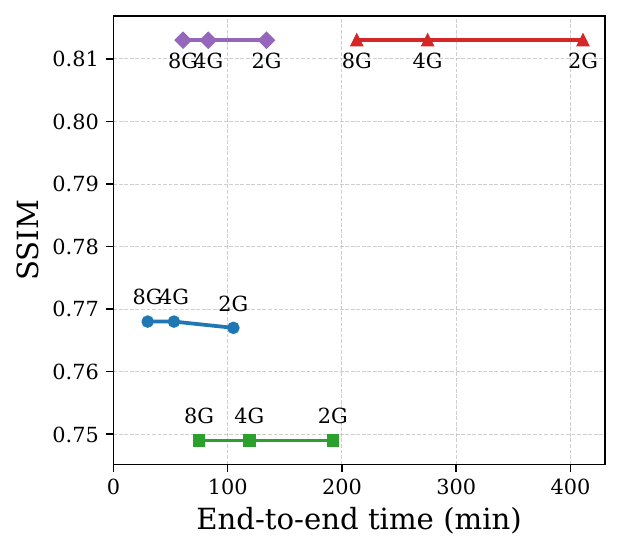} &
\rqfig{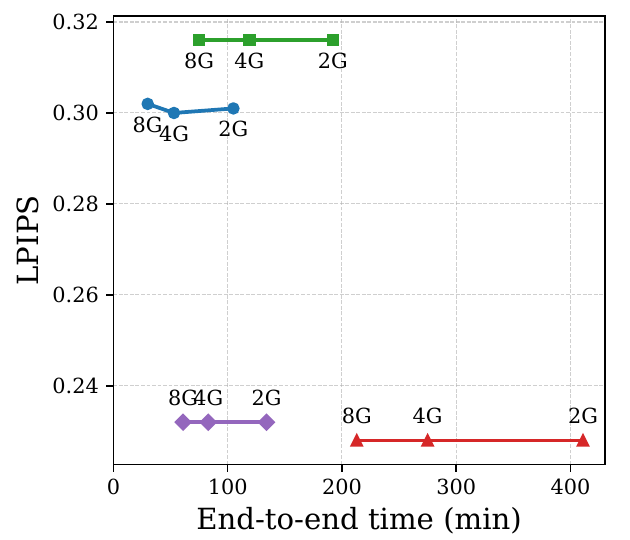} \\[2.5mm]
\rowname{Residence} &
\rqfig{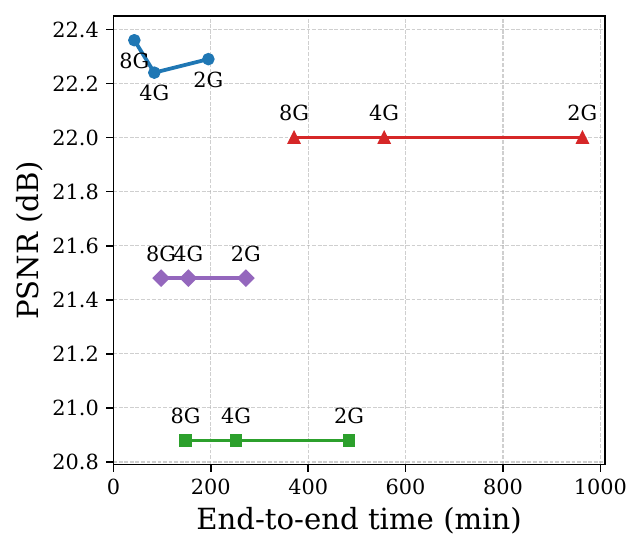} &
\rqfig{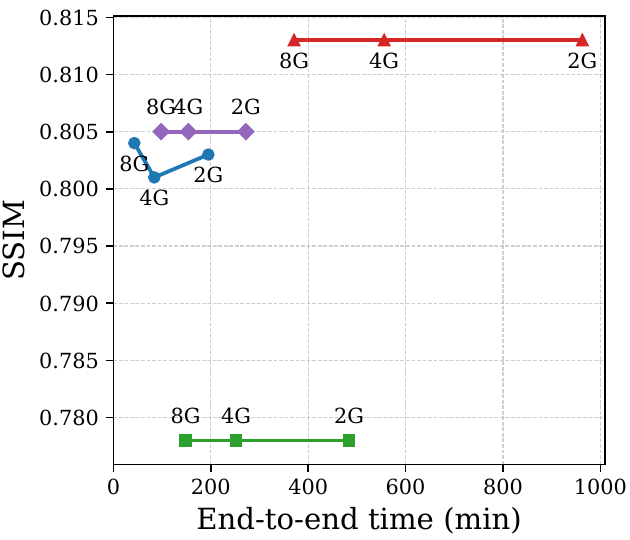} &
\rqfig{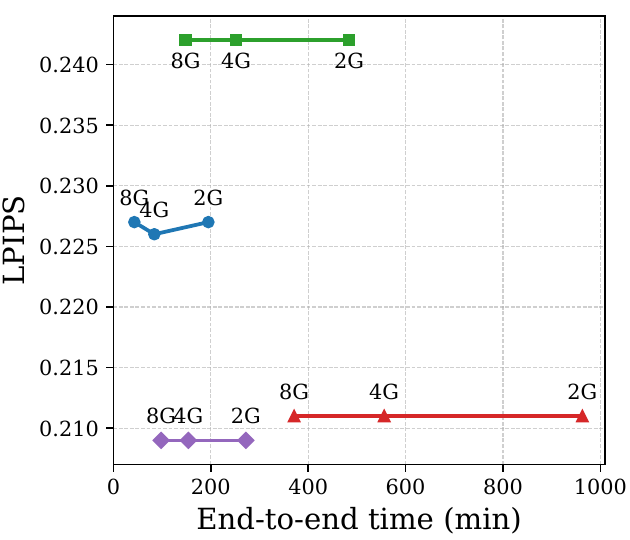} \\[2.5mm]
\rowname{Sci-Art} &
\rqfig{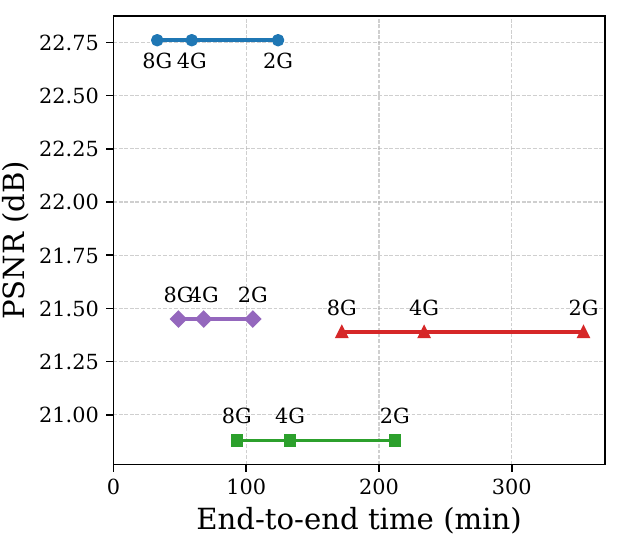} &
\rqfig{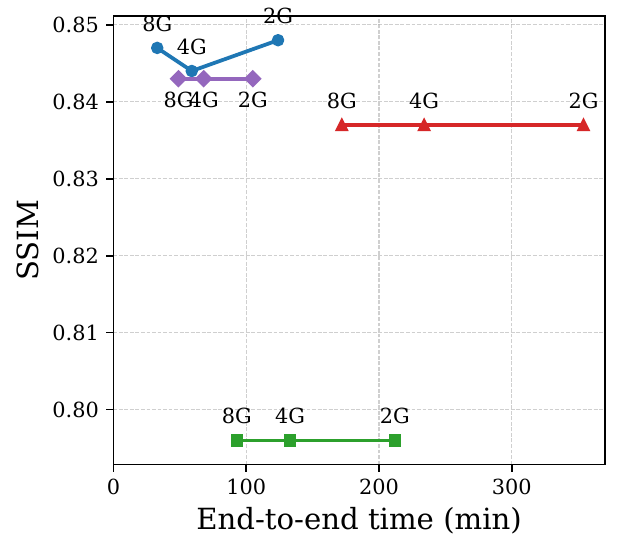} &
\rqfig{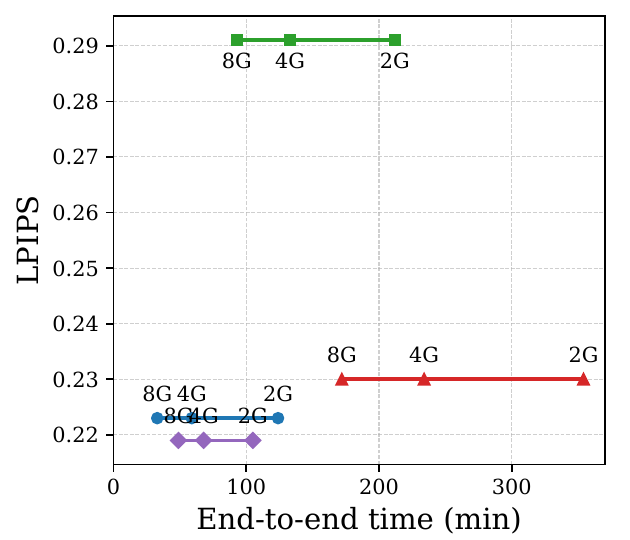} \\[2mm]
\end{tabular}
\vspace{-1mm}
\caption{\textbf{Quality vs.\ end-to-end runtime under the same GPU budget.}
Each row is a dataset; columns report PSNR, SSIM, and LPIPS against end-to-end
time. Per-method curves connect the 2, 4, and 8 GPU operating points. Up/left is
better for PSNR and SSIM; down/left is better for LPIPS.}
\label{fig:appendix_runtime_quality_tradeoff}
\end{figure*}

%% file: figure_tex/qualitative_full.tex
\begin{figure*}[!htbp]
\centering
\includegraphics[width=1.0\textwidth]{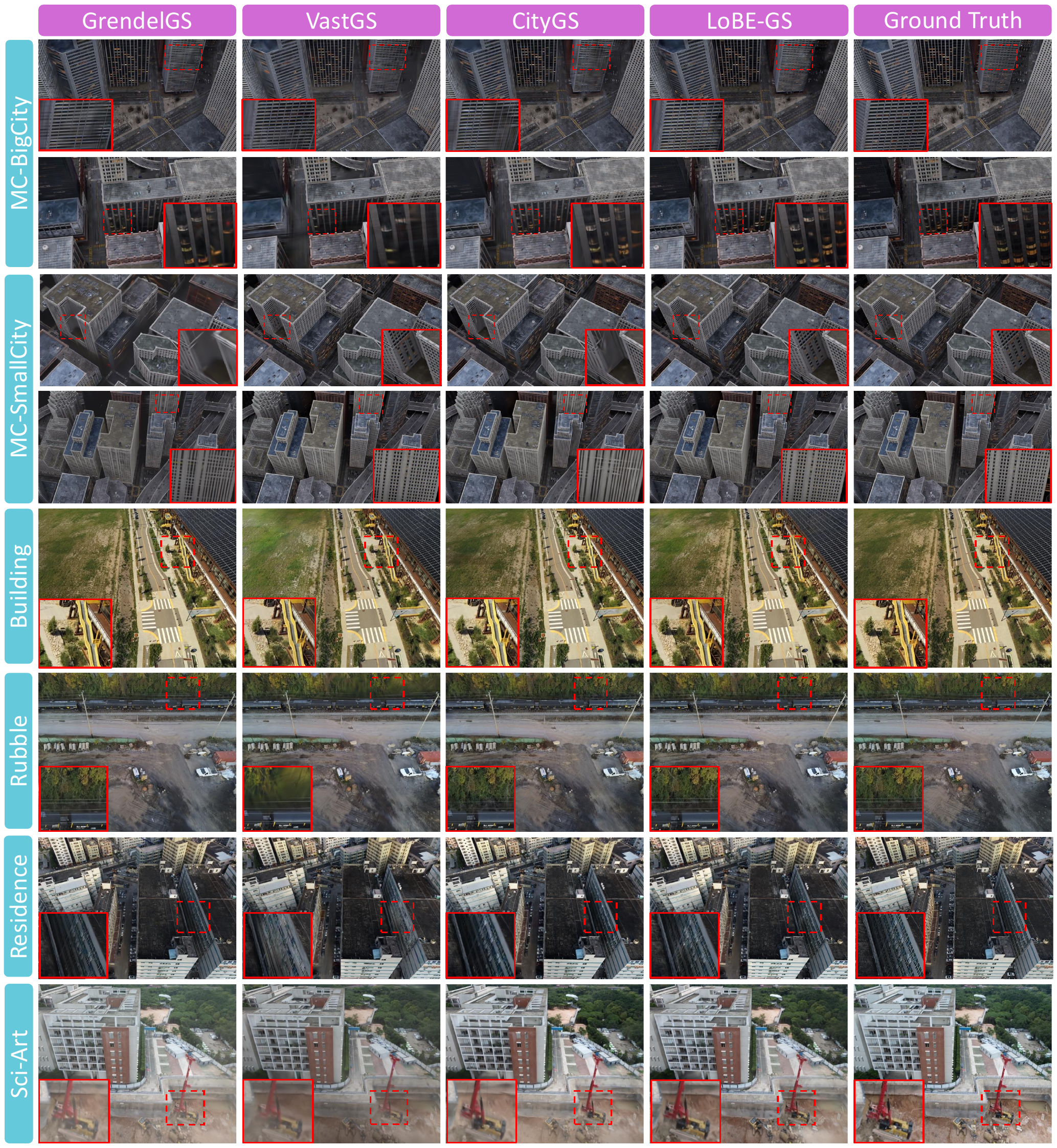}
\vspace{-5mm}
\caption{\textbf{Qualitative comparison on MatrixCity (MC), Mill19 and UrbanScene3D datasets.}}
\label{fig:qualitative_full}
\end{figure*}

%% file: figure_tex/qualitative_block_boundary.tex
\begin{figure*}[t]
  \centering
  \includegraphics[width=\textwidth]{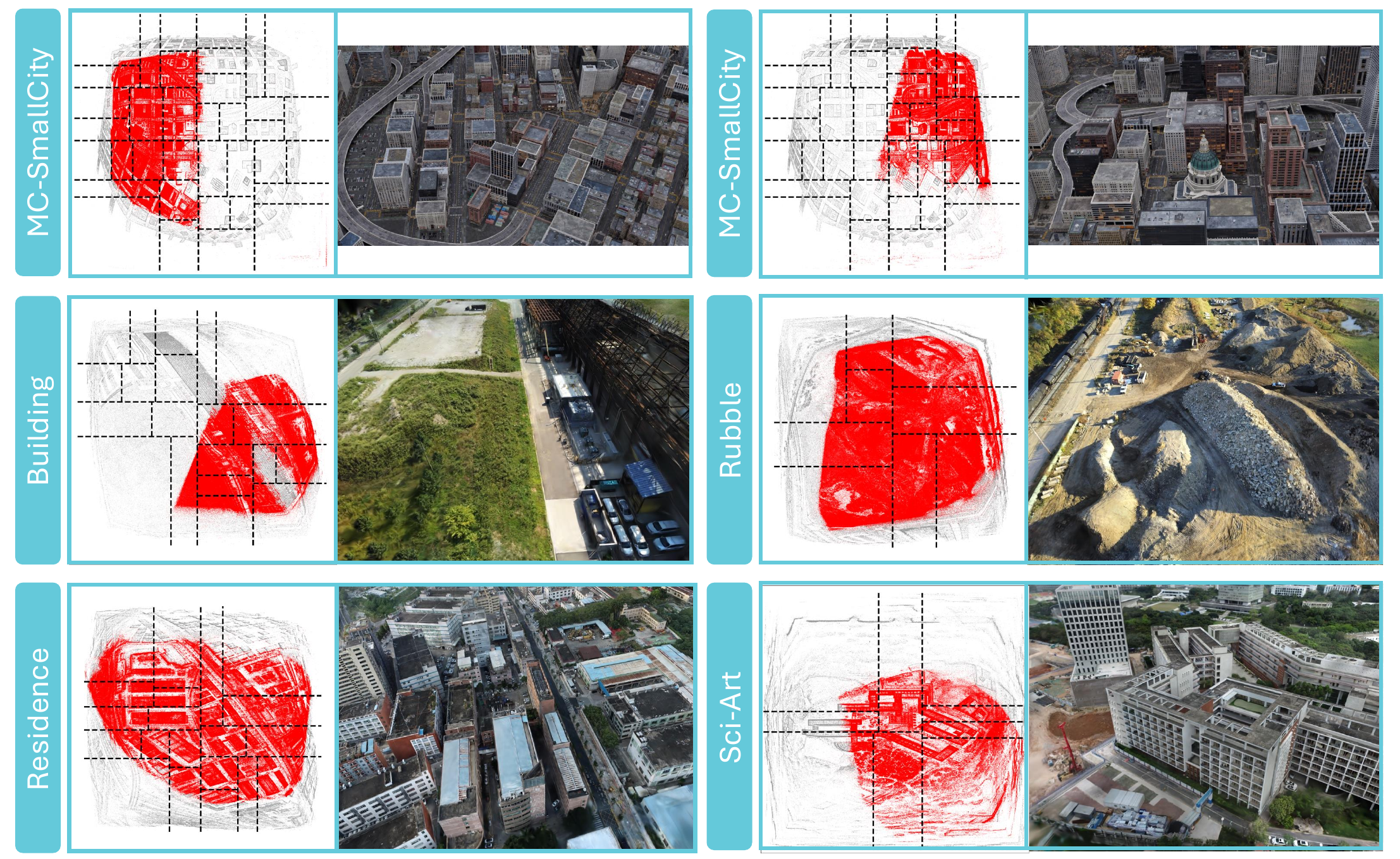}
  \caption{\textbf{Block boundary evaluation.} Top-view of the partitioned scene: gray points
  are Gaussians' positions, red points are visible Gaussians from the corresponding camera,
  and dashed lines mark the KD-tree block boundaries.}
  \label{fig:block_boundary}
\end{figure*}

%% file: table_tex/partition_strategies.tex
\begin{table}[t]
\centering
\small
\setlength{\tabcolsep}{5pt}
\begin{tabular}{@{}l ccc cc@{}}
\toprule
Method & PSNR & SSIM & LPIPS
& $T_{\mathrm{part}}$ & Max $T_{\mathrm{fine}}$ \\
\midrule
\multicolumn{6}{@{}l}{\textit{MatrixCity-SmallCity}} \\
VastGS  & 26.60 & 0.840 & 0.249 & 0:48 & 0:44 \\
CityGS  & \graybold{27.73} & 0.871 & 0.196 & 2:38 & 0:45 \\
LoBE-GS (Ours) & 27.69 & \graybold{0.875} & \graybold{0.185} & \graybold{0:08} & \graybold{0:27} \\
\midrule
\multicolumn{6}{@{}l}{\textit{Building}} \\
VastGS  & 21.65 & 0.765 & 0.265 & 0:05 & 0:35 \\
CityGS  & 22.01 & 0.782 & 0.249 & 0:23 & 0:40 \\
LoBE-GS (Ours) & \graybold{22.13} & \graybold{0.790} & \graybold{0.231} & \graybold{0:02} & \graybold{0:28} \\
\midrule
\multicolumn{6}{@{}l}{\textit{Rubble}} \\
VastGS  & 25.91 & 0.807 & 0.241 & 0:04 & 0:34 \\
CityGS  & 25.80 & 0.811 & 0.236 & 0:09 & 0:33 \\
LoBE-GS (Ours) & \graybold{25.93} & \graybold{0.813} & \graybold{0.232} & \graybold{0:01} & \graybold{0:25} \\
\midrule
\multicolumn{6}{@{}l}{\textit{Residence}} \\
VastGS  & 21.23 & 0.800 & 0.217 & 0:08 & 0:37 \\
CityGS  & \graybold{21.82} & \graybold{0.813} & 0.212 & 0:31 & 0:29 \\
LoBE-GS (Ours) & 21.48 & 0.805 & \graybold{0.209} & \graybold{0:03} & \graybold{0:26} \\
\midrule
\multicolumn{6}{@{}l}{\textit{Sci-Art}} \\
VastGS  & 21.37 & 0.834 & 0.234 & 0:25 & 0:21 \\
CityGS  & 21.19 & 0.835 & 0.235 & 0:08 & \graybold{0:20} \\
LoBE-GS (Ours) & \graybold{21.45} & \graybold{0.843} & \graybold{0.219} & \graybold{0:02} & \graybold{0:20} \\
\bottomrule
\end{tabular}
\caption{\graybold{Reconstruction quality and runtime across partition
strategies.} $T_{\mathrm{part}}$ is partitioning, Max $T_{\mathrm{fine}}$ the slowest block's fine-stage training. Times are $hh$:$mm$. Best per scene in \graylegend{bold}.}
\label{tbl:partition_strategies}
\end{table}

%% file: table_tex/cam_sel_aggreement.tex
\begin{table}[t]
\centering
\small
\setlength{\tabcolsep}{5pt}
\renewcommand{\arraystretch}{1.1}
\begin{tabular}{@{}l ccc @{\hspace{1.2em}} ccc@{}}
\toprule
& \multicolumn{3}{c}{LoBE-GS FCS (ours)}
& \multicolumn{3}{c}{\vastgs\ BBox} \\
\cmidrule(lr){2-4} \cmidrule(lr){5-7}
Dataset & Prec. & Rec. & F1 & Prec. & Rec. & F1 \\
\midrule
MC-SmallCity    & 0.641 & 0.970 & \textbf{0.772} & 0.340 & 0.934 & 0.499 \\
Building        & 0.792 & 0.914 & \textbf{0.848} & 0.677 & 0.895 & 0.771 \\
Rubble          & 0.783 & 0.965 & 0.865          & 0.856 & 0.913 & \textbf{0.883} \\
Residence       & 0.944 & 0.858 & \textbf{0.899} & 0.916 & 0.817 & 0.864 \\
Sci-Art         & 0.890 & 0.917 & \textbf{0.903} & 0.889 & 0.903 & 0.896 \\
\bottomrule
\end{tabular}
\caption{Agreement of each camera selection method with the \citygs\ SSIM
contribution reference, evaluated on a fixed coarse checkpoint. Precision and
recall are measured against the reference selection; higher values indicate closer agreement with \citygs\ SSIM-based camera selection. Bold marks the higher F1 per scene.}
\label{tbl:camera_agreement}
\end{table}

%% file: table_tex/camera_selection_sensitivity.tex
\begin{table}[!htbp]
\centering
\small
\begin{tabular}{@{}ccccccc@{}}
\toprule
\#Blocks & $\tau$ & PSNR & SSIM & LPIPS &  $T_{\text{partition}}$ \\
\midrule
\multirow{3}{*}{9}  & 0.05 & 27.30 & 0.858 & 0.220  & \multirow{3}{*}{00:04} \\
                    & 0.10 & 27.39 & 0.863 & 0.212  & \\
                    & 0.20 & \textbf{27.44} & \textbf{0.865} & \textbf{0.207} &  \\
\midrule
\multirow{3}{*}{16} & 0.05 & 27.61 & 0.869 & 0.200  & \multirow{3}{*}{00:05} \\
                    & 0.10 & \textbf{27.69} & 0.871 & 0.195 & \\
                    & 0.20 & 27.61 & \textbf{0.872} & \textbf{0.191} & \\
\midrule
\multirow{3}{*}{36} & 0.05 & \textbf{27.73} & 0.875 & 0.187  & \multirow{3}{*}{00:08} \\
                    & 0.10 & 27.69 & \textbf{0.875} & 0.185 & \\
                    & 0.20 & 27.57 & 0.874 & \textbf{0.180}  & \\
\bottomrule
\end{tabular}
\caption{Sensitivity to the selection threshold $\tau$ on MatrixCity-SmallCity. $T_{\text{partition}}$ is reported in $hh$:$mm$}
\label{tbl:tau}
\end{table}

%% file: figure_tex/partition_scalability.tex
\begin{figure}[t]
\centering
\begin{subfigure}[b]{\linewidth}
  \centering
  \includegraphics[width=\linewidth]{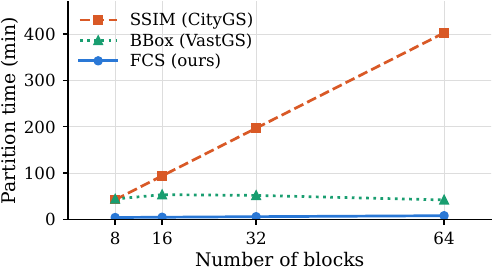}
  \caption{Varying block count, at $N_{\mathrm{cam}} = 5621$.}
  \label{fig:partition-blocks}
\end{subfigure}

\vspace{0.6em}

\begin{subfigure}[b]{\linewidth}
  \centering
  \includegraphics[width=\linewidth]{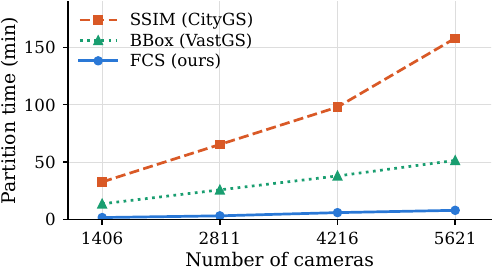}
  \caption{Varying camera count, at $N_{\mathrm{block}} = 36$.}
  \label{fig:partition-cams}
\end{subfigure}
\caption{Partition time of the three camera selection strategies. All three scale
linearly in camera count.}
\label{fig:partition-time}
\end{figure}

%% file: table_tex/coarse_model_sensitivity.tex
\begin{table}[H]
    \centering
    \setlength{\tabcolsep}{4pt}
    \renewcommand{\arraystretch}{1.1}
    \resizebox{1.0 \linewidth}{!}{%
    \begin{tabular}{l c | cccc | cccc}
        \toprule
        \multirow{2}{*}{Dataset} & \multirow{2}{*}{Coarse Iter.} &
        \multicolumn{4}{c|}{Coarse} &
        \multicolumn{4}{c}{Fine} \\
        & &
        PSNR & SSIM & LPIPS & \#GS &
        PSNR & SSIM & LPIPS & \#GS \\
        \midrule
        \multirow{3}{*}{SmallCity}
            & 10k & 23.32 & 0.668 & 0.491 & 5.6 & \textbf{27.80} & 0.874 & 0.186 & 27.9 \\
            & 20k & 24.55 & 0.730 & 0.414 & 8.8 & 27.71 & 0.873 & 0.186 & 28.5 \\
            & 30k & \textbf{25.27} & \textbf{0.765} & \textbf{0.368} & 13.6 & 27.69 & \textbf{0.875} & \textbf{0.185} & 28.9 \\
        \midrule
        \multirow{3}{*}{Building}
            & 10k & 18.93 & 0.628 & 0.396 & 4.5 & 21.89 & 0.788 & 0.234 & 14.0 \\
            & 20k & 19.86 & 0.682 & 0.348 & 6.7 & 21.97 & 0.789 & 0.233 & 13.6 \\
            & 30k & \textbf{20.42} & \textbf{0.709} & \textbf{0.327} & 9.5 & \textbf{22.13} & \textbf{0.790} & \textbf{0.231} & 13.9 \\
        \midrule
        \multirow{3}{*}{Rubble}
            & 10k & 22.56 & 0.640 & 0.423 & 3.7 & 25.94 & 0.809 & 0.239 & 9.2 \\
            & 20k & 24.04 & 0.721 & 0.344 & 5.9 & \textbf{26.10} & \textbf{0.813} & 0.233 & 9.3 \\
            & 30k & \textbf{24.82} & \textbf{0.753} & \textbf{0.310} & 7.8 & 25.93 & \textbf{0.813} & \textbf{0.232} & 9.2 \\
        \bottomrule
    \end{tabular}
    }
    \vspace{2mm}
    \caption{\textbf{Sensitivity of LoBE-GS to the quality of the coarse 3DGS prior.}
    We vary the number of coarse training iterations and report both coarse and final
    LoBE-GS reconstruction metrics on three datasets. \#GS is the number of Gaussians in millions.}
    \label{tbl:coarse_sensitivity}
\end{table}

%% file: figure_tex/load_balance_vis.tex
\begin{figure*}[!htbp]
    \centering
    \includegraphics[width=0.85
    \linewidth]{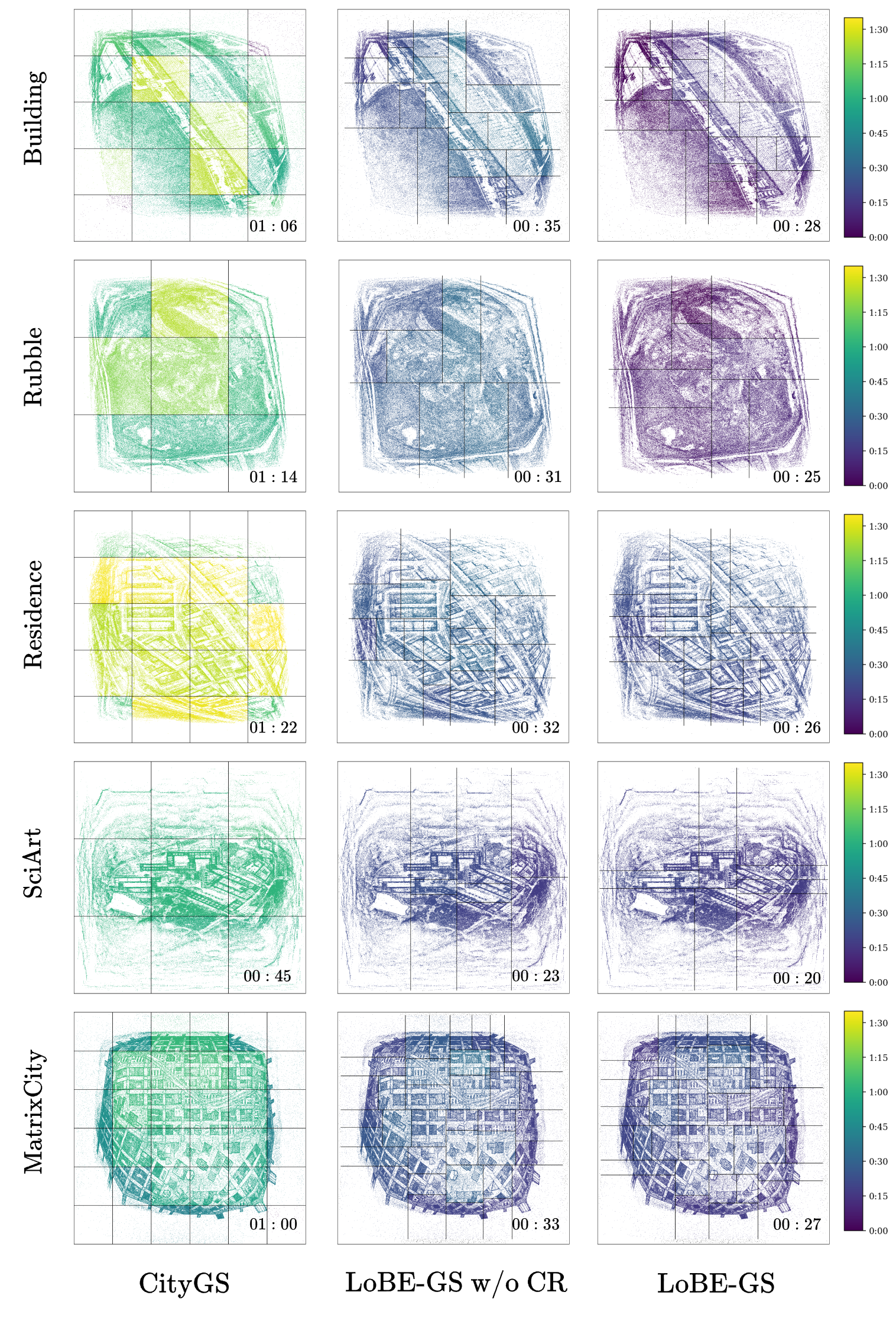}

    \caption{Visualization of the block partition and corresponding worst-block fine-stage runtime for \citygs{}, {\modelAbbrev} w/o cut refinement (CR), and full {\modelAbbrev} on five datasets: Building, Rubble, Residence, Sci-Art, and MatrixCity-SmallCity.}
    \label{fig:loadbalance_vis}
\end{figure*}

%% file: table_tex/cam_pose_bilateral_grid.tex
\begin{table*}[t]
\centering
\small
\setlength{\tabcolsep}{5pt}
\begin{tabular}{llccccccccc}
\toprule
Scene & Camera Pose & Bil.\ Grid & PSNR & SSIM & LPIPS & CC-PSNR & CC-SSIM & CC-LPIPS & \#GS (M)\\
\midrule
\multirow{3}{*}{Building}
& Original & \xmark & 22.13 & 0.790 & 0.231 & 22.97 & 0.786 & 0.235 & 13.9 \\
& COLMAP   & \xmark & \graybold{22.73} & \graybold{0.827} & 0.194 & \graybold{23.61} & \graybold{0.822} & 0.198 & 12.8 \\
& COLMAP   & \cmark & 20.27 & 0.808 & \graybold{0.193} & 22.95 & 0.818 & \graybold{0.191} & 14.3 \\
\midrule
\multirow{3}{*}{Rubble}
& Original & \xmark & 25.93 & 0.813 & 0.232 & 26.65 & 0.813 & 0.232 & 9.2 \\
& COLMAP   & \xmark & \graybold{27.45} & \graybold{0.888} & 0.164 & \graybold{28.10} & 0.887 & 0.164 & 11.9 \\
& COLMAP   & \cmark & 23.05 & 0.877 & \graybold{0.162} & 27.74 & \graybold{0.894} & \graybold{0.155} & 13.3 \\
\midrule
\multirow{3}{*}{Residence}
& Original & \xmark & 21.48 & 0.805 & 0.209 & 22.95 & 0.816 & 0.208 & 11.3 \\
& COLMAP   & \xmark & \graybold{22.14} & \graybold{0.841} & 0.174 & 23.21 & 0.839 & 0.184 & 12.1 \\
& COLMAP   & \cmark & 20.52 & 0.831 & \graybold{0.164} & \graybold{24.10} & \graybold{0.869} & \graybold{0.156} & 13.2 \\
\midrule
\multirow{3}{*}{SciArt}
& Original & \xmark & \graybold{21.45} & 0.843 & 0.219 & 24.54 & 0.852 & 0.218 & 4.5 \\
& COLMAP   & \xmark & 21.44 & 0.841 & 0.222 & 24.44 & 0.849 & 0.221 & 4.6 \\
& COLMAP   & \cmark & 19.77 & \graybold{0.844} & \graybold{0.193} & \graybold{25.86} & \graybold{0.876} & \graybold{0.185} & 5.2 \\
\bottomrule
\end{tabular}
\caption{\textbf{Quantitative result under different camera pose
sources and appearance modeling.} \emph{Pose}
indicates whether the camera poses distributed with the dataset
(\emph{Original}) or our re-estimated poses (\emph{COLMAP}) are used;
\emph{Bil.\ Grid} indicates the bilateral grid appearance module. CC denotes
color-corrected metrics; best value per scene and metric in
\graylegend{bold}.}
\label{tab:appendix_pose_appearance}
\end{table*}

%% file: figure_tex/qualitative_colmap_bilateral_grid.tex
\begin{figure*}[t]
\centering
\includegraphics[width=\textwidth]{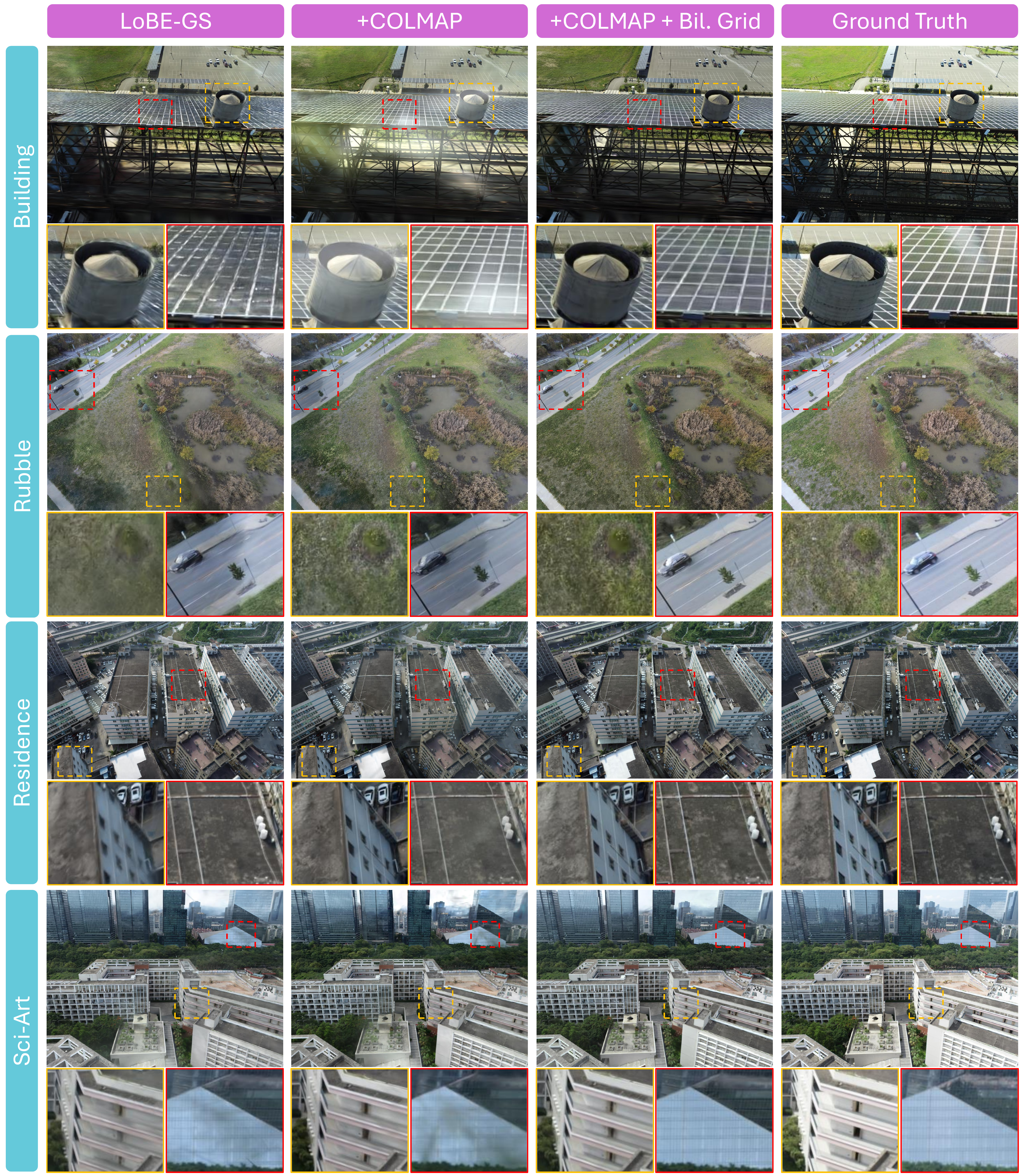}
\caption{\textbf{Qualitative comparison under different camera pose
sources and appearance modeling.} Insets show
magnified crops of the boxed regions. Renders using the original camera poses
exhibit blurry artifacts on high-frequency structure, while the bilateral grid reduces
low-frequency color blotching.}
\label{fig:qualitative}
\end{figure*}

%% file: table_tex/algo.tex
\begin{algorithm}
\footnotesize
\DontPrintSemicolon
\SetKwFunction{FPartition}{PartitionScene}
\SetKwFunction{FRefine}{RefineCutline}
\SetKwFunction{FGetWorkload}{ComputeWorkload}
\SetKwInOut{Input}{Input}\SetKwInOut{Output}{Output}
\Input{Scene AABB $\gB^{(0)}$, camera set $\gC^{(0)}$,\\
       \#blocks $M$, coarse Gaussian centers $\{g_i\}$}
\Output{$M$ leaf nodes $\{\gB^{(b)}, \gC^{(b)}\}_{b=1}^{M}$}
\BlankLine
Initialize an empty priority queue $PQ$ \\
$PQ.\mathrm{push}\big(|\gC^{(0)}|,\ \mathrm{root}\big)$\;
\BlankLine
\While{$|PQ| < M$}{
    Pop node $(\gB^{\mathrm{parent}}, \gC^{\mathrm{parent}})$ with largest workload $L(\gB^{\mathrm{parent}})=|\gC^{\mathrm{parent}}|$\;
    \BlankLine
    \tcp{Select split axis}
    $a \leftarrow \underset{d \in \{x,y\}}{\arg\max}\ \mathrm{ext}_d\big(\gB^{\mathrm{parent}}\big)$\;
    \tcp{$s_a(p)$: $p$-th quantile of $\Pi_a(\{g_i \in \gB^{\mathrm{parent}}\})$}
    \BlankLine
    \tcp{Cutline refinement in percentile space}
    $p^{*}\leftarrow 0.5$ \tcp*{median split (Sec.~3.1)}
    $p_{\min}\leftarrow 0.1$,\quad $p_{\max}\leftarrow 0.9$\;
    $N_{\mathrm{steps}}\leftarrow 3$,\quad $J_{\mathrm{best}}\leftarrow \infty$\;
    \For{$iter \leftarrow 1$ \KwTo $N_{\mathrm{steps}}$}{
        Split $\gB^{\mathrm{parent}}$ at $s_a(p^{*})$ into $\gB_L,\,\gB_R$\;
        Assign $\gC_L,\gC_R \subseteq \gC^{\mathrm{parent}}$ via FCS\ (main paper, Sec.~3.2)\;
        $J \leftarrow \max\big(|\gC_L|,\,|\gC_R|\big)$ \tcp*{Eq.~(1)}
        \If{$J < J_{\mathrm{best}}$}{
            $J_{\mathrm{best}} \leftarrow J$\;
            $(\hat{\gB}_L,\hat{\gC}_L,\hat{\gB}_R,\hat{\gC}_R) \leftarrow (\gB_L,\gC_L,\gB_R,\gC_R)$\;
        }
        \If{$|\gC_L| > |\gC_R|$}{
            $p_{\max}\leftarrow p^{*}$;\ \ $p^{*}\leftarrow (p_{\min}+p^{*})/2$\;
        }
        \ElseIf{$|\gC_R| > |\gC_L|$}{
            $p_{\min}\leftarrow p^{*}$;\ \ $p^{*}\leftarrow (p^{*}+p_{\max})/2$\;
        }
        \Else{
            \textbf{break}\;
        }
    }
    \BlankLine
    $PQ.\mathrm{push}\big(|\hat{\gC}_L|,\ (\hat{\gB}_L, \hat{\gC}_L)\big)$\;
    $PQ.\mathrm{push}\big(|\hat{\gC}_R|,\ (\hat{\gB}_R, \hat{\gC}_R)\big)$\;
}
\Return all nodes in $PQ$\;
\caption{Load-balanced KD-tree Scene \\ Partitioning}
\label{alg:kd_tree_partition}
\end{algorithm}